\documentclass[sigconf]{acmart}
\acmSubmissionID{983}

\usepackage{graphicx, amsmath, amssymb, caption, subcaption, multirow, overpic, textpos}
\usepackage{tabularx}
\usepackage{colortbl}

\newlength\savewidth

\renewcommand{\paragraph}[1]{\vspace{1.25mm}\noindent\textbf{#1}}

\newcolumntype{x}[1]{>{\centering\arraybackslash}p{#1pt}}
\newcolumntype{y}[1]{>{\raggedright\arraybackslash}p{#1pt}}
\newcolumntype{z}[1]{>{\raggedleft\arraybackslash}p{#1pt}}

\newcommand{\app}{\raise.17ex\hbox{$\scriptstyle\sim$}}

\definecolor{deemph}{gray}{0.6}

\definecolor{baselinecolor}{gray}{.9}

\newcommand{\gr}{\rowcolor[gray]{.95}}

\AtBeginDocument{%
  \providecommand\BibTeX{{%
    \normalfont B\kern-0.5em{\scshape i\kern-0.25em b}\kern-0.8em\TeX}}}

\setcopyright{acmcopyright}
\copyrightyear{2018}
\acmYear{2018}
\acmDOI{XXXXXXX.XXXXXXX}

\acmConference[Conference acronym 'XX]{Make sure to enter the correct
  conference title from your rights confirmation emai}{June 03--05,
  2018}{Woodstock, NY}
\acmPrice{15.00}
\acmISBN{978-1-4503-XXXX-X/18/06}

\renewcommand\footnotetextcopyrightpermission[1]{}
\begin{document}

\title{Towards Efficient Single Image Dehazing and Desnowing}

\author{Tian Ye}
\authornote{Equal contribution.\\ $^\dag$Corresponding author.}
\email{201921114031@jmu.edu.cn}
\affiliation{%
  \institution{School of Ocean Information Engineering, Jimei University}
  %\streetaddress{P.O. Box 1212}
  \city{Xiamen}
  \state{Fujian}
  \country{China}
}

\author{Sixiang Chen}
\authornotemark[1]
\email{201921114013@jmu.edu.cn}
\affiliation{%
  \institution{School of Ocean Information Engineering, Jimei University}
  %\streetaddress{P.O. Box 1212}
  \city{Xiamen}
  \state{Fujian}
  \country{China}
}

\author{Yun Liu}
\email{yunliu@swu.edu.cn}
\authornotemark[1]
\affiliation{%
  \institution{College of Artificial Intelligence, Southwest University}
  %\streetaddress{P.O. Box 1212}
  \city{Chonqing}
  \country{China}
}

\author{Erkang Chen$^\dag$}
\email{ekchen@jmu.edu.cn}
\affiliation{%
  \institution{School of Ocean Information Engineering, Jimei University}
  %\streetaddress{P.O. Box 1212}
  \city{Xiamen}
  \state{Fujian}
  \country{China}
}

\author{Yuche Li}
\email{liyuche\_cq@163.com}
\affiliation{%
  \institution{College of Geosciences, China University of Petroleum}
  %\streetaddress{P.O. Box 1212}
  \city{Beijing}
  \country{China}
}



\begin{abstract}
Removing adverse weather conditions like rain, fog, and
snow from images is a challenging problem. Although the current recovery algorithms targeting a specific condition have made impressive progress, it is not flexible enough to deal with various degradation types. We propose an efficient and compact image restoration network named DAN-Net (Degradation-Adaptive Neural Network) to address this problem, which consists of multiple compact expert networks with one adaptive gated neural. A single expert network efficiently addresses specific degradation in nasty winter scenes relying on the compact architecture and three novel components. Based on the Mixture of Experts strategy, DAN-Net captures degradation information from each input image to adaptively modulate the outputs of task-specific expert networks to remove various adverse winter weather conditions. Specifically, it adopts a lightweight Adaptive Gated Neural Network to estimate gated attention maps of the input image, while different task-specific experts with the same topology are jointly dispatched to process the degraded image. Such novel image restoration pipeline handles different types of severe weather scenes effectively and efficiently. It also enjoys the benefit of coordinate boosting in which the whole network outperforms each expert trained without coordination. 

Extensive experiments demonstrate that the presented manner outperforms the state-of-the-art single-task methods on image quality and has better inference efficiency. Furthermore, we have collected the first real-world winter scenes dataset to evaluate winter image restoration methods, which contains various hazy and snowy images snapped in winter. 
Both the dataset and source code will be publicly available.
%It will be released with source code after our work has been accepted.

  \end{abstract}




\begin{teaserfigure}
    \centering
    \includegraphics[width=\textwidth]{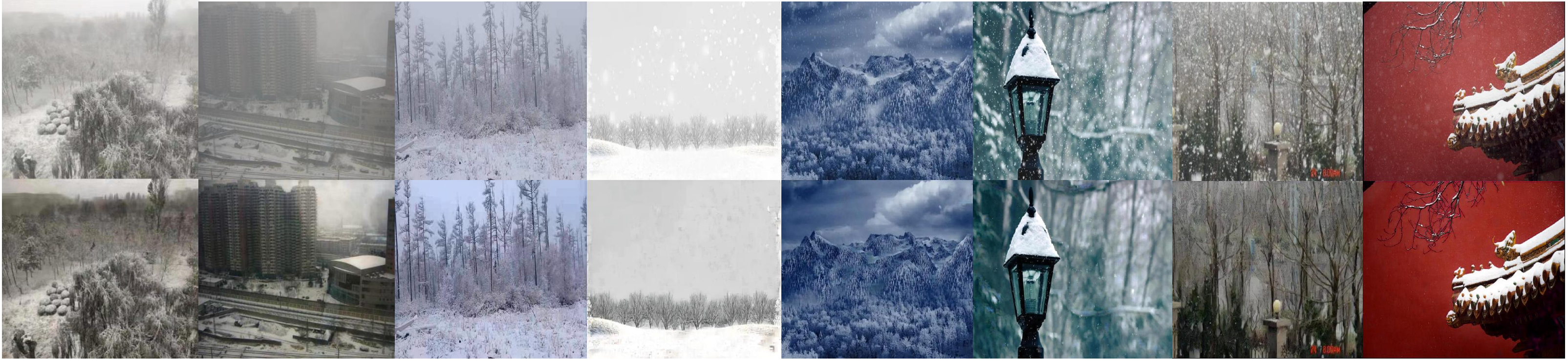}
    \caption{Realistic winter images (top) from the RWSD dataset created by this work and corresponding restoration results (bottom) using the proposed Degradation-Adaptive Neural Network. }
    \label{fig:realisticrestoration}
\end{teaserfigure}

%(b)The PSNR vs. Parameters of state-of-the-art deep learning dehazing methods and our methods on the haze4k ~\cite{liu2021synthetic} dataset. (c)The PSNR vs. Parameters of state-of-the-art deep learning desnowing methods and our methods on the CSD ~\cite{chen2021all} dataset. Note that our universal restoration method achieves the best performance-parameters trade-off on both task.

\maketitle

\section{Introduction}
 In nasty weather scenes, a series of high-level vision tasks like object recognition~\cite{redmon2016you,shafiee2017fast,szegedy2013deep}, person re-identification~\cite{hirzer2011person,li2018harmonious,wu2019deep,zheng2011person} urgently need trustworthy image restoration. %Removing adverse weather conditions like rain, fog, and snow from images is a fundamental problem for practical applications.
 
Recently, various end-to-end CNN-based methods ~\cite{li2020all,ffa-net,wu2021contrastive,liu2021synthetic,msbdn,aod} have been proposed to simplify the image restoration problem by directly learning the mapping of degraded-to-clear, it achieves impressive performance. However, due to the studying difficulty of universal methods, most previous image restoration methods only handle single degeneration by lousy weather, e.g., hazy, rainy, and snowy, thus resulting in their methods only performing well on the typical scene. In real-world scenes, the simultaneous existence of multiple types of degradation is more common than a single degradation. People often meet the complex degenerated scenes by bad weather; for instance, the authentic snowy images are usually degraded with haze and snowy or rainy streaks.

In this work, we focus on developing an efficient and satisfactory solution for image restoration in winter scenes. Complex winter scenes contain diverse snow streaks, snowflakes, and the uneven haze effect, resulting in a more challenging and less-touch vision task than rainy scenes. 

In daily winter scenes, where uneven haze often occurs with diverse snowy degradation, the widely used haze imaging model is quite different from the snow imaging model. The haze-imaging can be modeled as the following formation:
\begin{equation}
    \mathcal{I}(x) = \mathcal{J}(x)t(x) + \mathcal{A}(x)(1-t(x))
\end{equation}
where $t(x)$ is the transmission map of hazy image $\mathcal{I}(x)$, and $\mathcal{A}(x)$ is the global atmospheric light. Single image dehazing methods~\cite{he2010single,ffa-net,msbdn,aod} aim to recover the clean image $\mathcal{J}(x)$ from $\mathcal{I}(x)$.

According to the previous works~~\cite{chen2020jstasr,chen2021all}, the formation of snow is modeled as:
\begin{equation}
\label{eq:I(x)}
    \mathcal{I}(x) = \mathcal{K}(x)\mathcal{T}(x)+\mathcal{A}(x)(1-\mathcal{T}(x))
\end{equation}
where $\mathcal{K}(x) = \mathcal{J}(x)(1-\mathcal{Z}(x)\mathcal{R}(x))+\mathcal{C}(x)\mathcal{Z}(x)\mathcal{R}(x)$, $\mathcal{I}(x)$ represents the snowy image, $\mathcal{K}(x)$ denotes the veiling-free snowy image, $\mathcal{T}(x)$ is the media transmission map, $\mathcal{A}(x)$ is the atmospheric light and $\mathcal{J}(x)$ is the snow-free image. $\mathcal{C}(x)$ and $\mathcal{Z}(x)$ are the chromatic aberration map for snow images and the snow mask, respectively. $\mathcal{R}(x)$ is the binary mask, presenting the snow location information.

Obviously, the incompatibility of physical models implies a considerable challenge to remove haze and snow for a single compact network coordinately. The diverse degradation scale of the snowy and uneven density of natural haze leads to a high threshold for efficient implementation.
 
Based on the above analysis, most existing methods have been designed to overcome just one type of degradation. Recently, a CNN-based universal method,\textit{i.e.},All-in-One~\cite{li2020all}, is proposed to remove all weather conditions at once. However, this method is limited in the real-world practical application due to the heavy amount of parameters, and meanwhile it ignores the implicit useful knowledge in different types of degradation to boost the performance of the model. Therefore, there is still room for further improvement.

In order to achieve efficient and practical real-world winter image restoration, we propose an efficient Degradation-Adaptive Neural (DAN) Network, which improves the restoration ability via the MoE (Mixture of Experts) strategy. Specifically, we carefully design high-performance components to construct the specific-task expert network: Multi-branch Spectral Transform Block based on Fast Fourier Convolution and multi-scale filter to address the diverse scale degradation, Dual-Pool Attention Module to refine the uneven features of real-world degradation, and Cross-Layer Activation Gated Module to optimize information-flow of features with different scales. To our surprise, both the lightweight dehazing and desnowing expert networks can achieve state-of-the-art performance and best performance-params trade-off compared with existing single task methods. %this success as a low-hang fruit inspired us to further utilize the above expert networks for a universal image restoration solution in winter scenes. 
%avoid the forgetting problem of neural network and 
How to skillfully combine the expert network with different task knowledge is a desirable topic for us. We rethink this fundamental topic and propose the light-weight Adaptive Gated Neural to learn how to fully utilize the inner knowledge of expert networks for better restoration performance. Furthermore, firstly, we pre-train two expert networks for dehazing and desnowing, respectively. Then, the Adaptive Gated Neural is presented to estimate gated attention maps for different expert networks to modulate the contributions. The Degradation-Adaptive Neural Network as an efficient and effective image restoration solution addresses diverse degradation in winter scenes, the complex, uneven hazy effects, and snow particles, streaks, and patches.

Our main contributions are summarized as follows:

\begin{itemize}

\item We propose three high-performance modules as fundamental components to construct a compact network as the task-specific expert network, which achieves state-of-the-art performance on image dehazing and desnowing.

\item The proposed Adaptive Gated Neural as an effective degradation-adaptive guider successfully modulates the contributions of task-specific expert networks.

\item We conduct extensive experiments and ablation studies to evaluate the presented method. The results demonstrate that our DAN-Net performs favorably against state-of-the-art dehazing or desnowing methods. To the best of our knowledge, it's the first time a universal solution achieves the best performance on two different tasks.

\item We have noticed the lack of natural winter scenes dataset that contains both hazy and snowy degradation, which is not conducive to the practical application of current networks training on synthetic domain. We collected and sorted the first real-world winter scenes dataset, namely RWSD, which will be open source with the proposed networks later.

%satisfactory image quality on both synthetic and real-world degraded images of winter scenes efficiently.

\end{itemize}

\section{Related Work}
\subsection{Single Image Dehazing}
\textbf{Prior-based Dehazing Methods.} Image dehazing was traditional image processing methods based on physical models when deep learning was not yet popular, which mostly utilized various priors to obtain the prerequisites of dehazing, such as dark channel prior \cite{he2010single}, color lines prior \cite{fattal2014dehazing}, haze-lines \cite{berman2017air}, color attenuation prior \cite{zhu2015fast}, region lines prior \cite{ju2021idrlp} and non-local prior \cite{berman2016non}. The most well-known DCP used the dark channel prior to estimate the transmission map whose principle is based on the assumption that the local patch of the haze-free image is close to zero in the lowest pixel of the three channels. Following the idea of DCP, there are many improved versions that have achieved amazing effects in dehazing \cite{xu2012fast,wang2015single,xie2010improved}. Though prior-based dehazing methods performed nontrivially, which often can't rectify the weaknesses of limiting by sophisticated scenes. \\
\textbf{Learning-based Dehazing Methods.} Numerous learning-based methods have certain limitations. With the development of convolutional neural networks, data-driven deep learning methods are widely used in the field of image dehazing \cite{li2017all,cai2016dehazenet,ffa-net,msbdn,griddehazenet,shao2020domain,liu2021synthetic}. DehazeNet \cite{cai2016dehazenet} utilized an end-to-end network to learn the transmission graph in the physical model to dehaze. FFA-Net \cite{ffa-net} designed an attention network for feature map fusion, which combined pixel attention and channel attention for image dehazing. Dong $et$ $al.$ \cite{msbdn} presented a multi-scale boosted dehazing network with dense feature fusion based on the U-Net backbone which performed amazingly. For domain gap between real and synthetic condition, Shao $et$ $al.$ \cite{shao2020domain} developed a bidirectional translation module and two image dehazing modules to incorporate the real hazy image into the dehazing training via exploiting the properties of the clear image for overcoming this issue.

\subsection{Single Image Desnowing}

\textbf{Prior-based Desnowing Methods.} Early approaches~~\cite{xu2012An,xu2012removing,rajderkar2013removing,zheng2013single,pei2014Removing,wang2017A} utilize hand-crafted priors to achieve desnowing from a single snowy image. Pei \emph{et al}.~~\cite{pei2014Removing} make use of features on saturation and visibility to remove the snowflakes. A guidance image method is proposed by Xu \emph{et al}.~~\cite{xu2012An} to remove snow from a single image. Based on the difference between background edges and rain streaks, Zheng \emph{et al}.~~\cite{zheng2013single} utilize multi-guided filter to extract the features for splitting the snowy component from the background. Wang \emph{et al}.~~\cite{wang2017A} develop a 3-layer hierarchical scheme for desnowing by leveraging image decomposition and dictionary learning.\\
\textbf{Learning-based Desnowing Methods.} For learning-based desnowing algorithms in the single image~~\cite{liu2018desnownet,engin2018cycle,li2020all,chen2020jstasr,chen2021all}, the first presented desnowing network called DesnowNet~~\cite{liu2018desnownet} deals with the removal of translucent and opaque snow particles in sequence based on a multi-stage architecture. Li~\emph{et al}.~~\cite{li2020all} develop an all-in-one network consisting of multiple task-specific encoders that can simultaneously handle multiple types of bad weather scenarios, such as rain, snow, haze and raindrops. JSTASR~~\cite{chen2020jstasr} propose a novel snow formation model and size-aware snow network for single image desnowing which takes the veiling effect and variety of snow particles into consideration. HDCW-Net~~\cite{chen2021all} performs snow removal by embedding the dual-tree wavelet transform into the architecture of the network and designing prior-based supervised loss called the contradict channel on the basis of the differences of snow and clean images. In the meantime, two large-scale snow datasets named snow removal in realistic scenario (SRRS) and comprehensive snow dataset (CSD) are respectively proposed in~~\cite{chen2020jstasr} and~~\cite{chen2021all} for promoting the development of learning-based image desnowing methods.

% This section presents the details of our Spatial Factor self-disillation framework. First, we provide details of the detail of student network and the proposed Factor Self-Distillation method. Then, we describe the overview of the Multi-Scale Spectral Transform Network and the all blocks in the proposed MST-Net.

\section{Task-Specific Expert Network}
In this section, we firstly present the details of high-performance conv-based components we designed for the better performance of expert network. We mainly utilize the Dual-pool attention module, Multi-branch Spectral Transform block and Cross-Layer Activation Gated module to structure efficient expert network. According to the previous image restoration methods~\cite{ffa-net,msbdn,liu2021synthetic,chen2020jstasr}, impressive performance is often achieved by simply stacking the components repeatedly, which we believe is a design idea with lower benefits. We deliberately avoided this solution and achieved excellent performance-param trade-off through careful design in each layer and as few components as possible. 

\begin{figure*}[t]
    \centering
    \includegraphics[width=\textwidth]{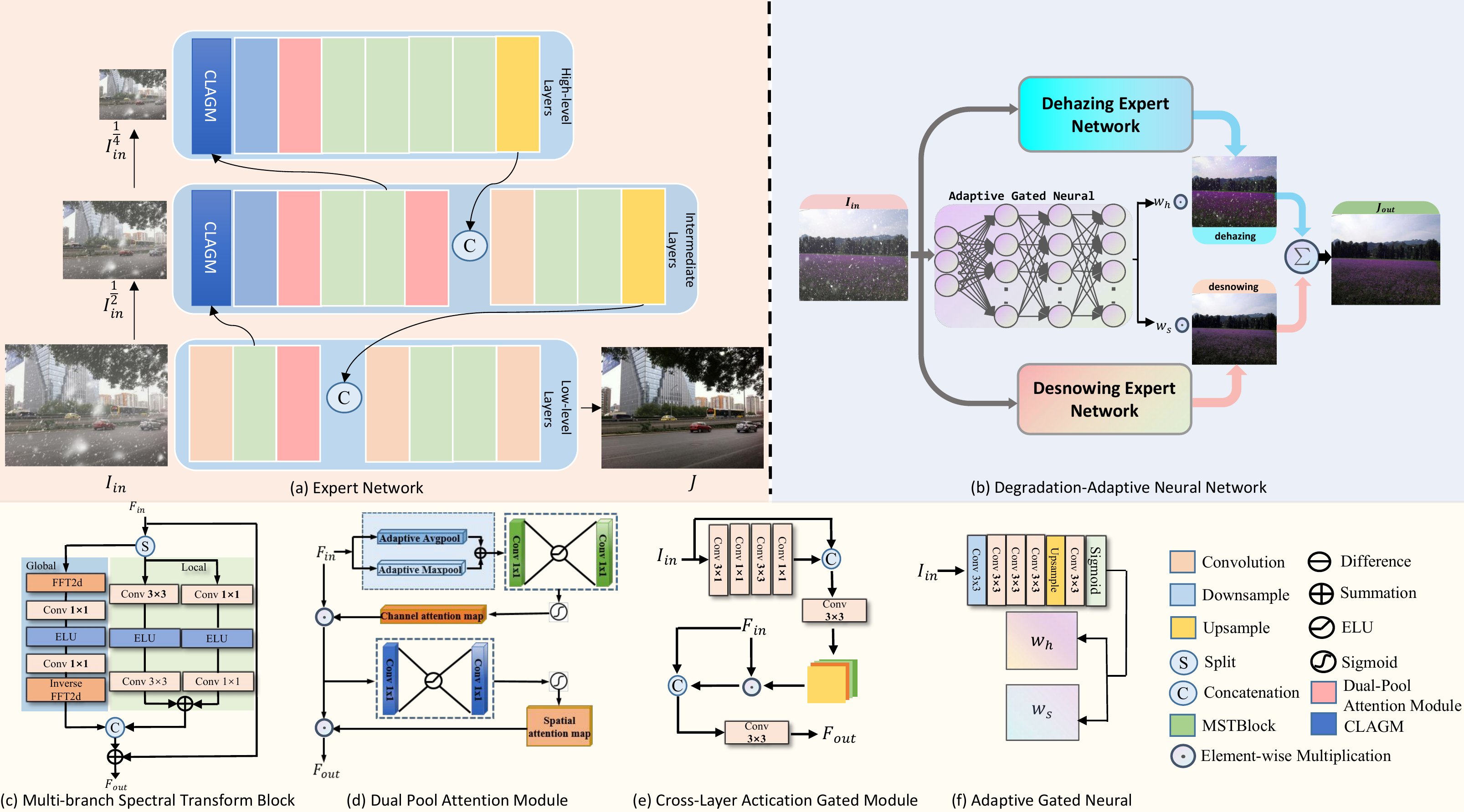}
    \caption{\small{The schematic illustration of the proposed task-specific Expert Network and the DAN-Net. We first provide the well-trained Expert Network for image dehazing or desnowing and utilize the Adaptive Gated Neural module to dynamically control the activation of expert networks for coordinated removing haze and snow from a winter image. We give details of the structure and configurations of the proposed DAN-Net in Sections 3 and 4.}}
    \label{fig:overviewofarchitecture}
\end{figure*}

\paragraph{\textbf{Dual-pool Attention Module.}}
More and more works notice that the importance of attention module for the represent ability of network in image restoration task. We introduce the dual-pool attention module into our network for better performance. As shwon in Fig.\ref{fig:overviewofarchitecture}(d), firstly, we utilize the AvgPool and Maxpool to sample the incoming features for the generation of channel dimension weights:
\begin{equation}
\begin{aligned}
  \text{AvgPool}: \mathbb{R}^{H \times W \times C} \rightarrow \mathbb{R}_{avg}^{1 \times 1 \times C} \\
  \text{MaxPool}: \mathbb{R}^{H \times W \times C} \rightarrow \mathbb{R}_{max}^{1 \times 1 \times C} \\
\end{aligned}
\end{equation}
And we add the sampled features, utilize the convolution block to generate the attention weights of channel dimension:
\begin{equation}
\text { Conv } \circ \text { ELU } \circ \text { Conv with } 1 \times 1 : \mathbb{R}_{avg}^{1 \times 1 \times C}+\mathbb{R}_{max}^{1 \times 1 \times C} \rightarrow \mathbb{R}^{1 \times 1 \times C}
\end{equation}
The $Sigmoid$ function is used to keep the boundary of final attention weights:
\begin{equation}
    \text{Sigmoid} : \mathbb{R}^{1 \times 1 \times C} \rightarrow \mathbb{R}_{c}^{1 \times 1 \times C},
\end{equation}
The attention mechanism in spatial dimension is also necessary for refining the features. The plain convolution block can generate the attention map in 2D dimension:
\begin{equation}
\operatorname{Conv} \circ \operatorname{ELU} \circ \operatorname{Conv} \text{with} 1 \times 1: \mathbb{R}^{H \times W \times C } \rightarrow \mathbb{R}^{H \times W \times \frac{C}{r}} \rightarrow  \mathbb{R}^{H \times W \times 1},
\end{equation}
where $r$ is the scale-factor, we set it as $4$ for all experiments. The $Sigmoid$ function constrains the final spatial attention map:
\begin{equation}
    \text{Sigmoid} : \mathbb{R}^{H \times W \times 1} \rightarrow \mathbb{R}_{s}^{H \times W \times 1},
\end{equation}

\paragraph{\textbf{Multi-branch Spectral Transform Block.}}

Considering the uneven distribution of features in channel dimension in neural networks, we assume that the information from a part of the channel in incoming features can help network to establish global dependencies by spectral transformation. Consequently, Multi-branch Spectral Transform (\textbf{MST}) block we designed splits the incoming features $\mathbb{R}_{in}$ firstly,
\begin{equation}
\mathbb{R}_{in}^{ H \times W \times C} \rightarrow \mathbb{R}_{g}^{ H \times W \times \frac{C}{2} },\mathbb{R}_{l}^{ H \times W \times \frac{C}{2}}    
\end{equation}
and the global branch of MST performs Fast Fourier Convolution (\textbf{FFC}) on the $\mathbb{R}_{g}^{H \times W \times \frac{C}{2} }$. As shown in Fig~\ref{fig:overviewofarchitecture}.(c), the global branch utilizes real FFT to account for global context. This designed parallel multi-branch combines different scale operation in single block, effectively capturing local information and global dependencies simultaneously and avoiding heavy computational burden for a lightweight network. Specifically, the global branch makes following steps, applies 2D Real FFT to the input features firstly:
\begin{equation}
\text { Real FFT 2d} : \mathbb{R}_{g}^{H \times W \times \frac{c}{2}} \rightarrow \mathbb{C}^{H \times \frac{W}{2} \times \frac{C}{2}}
\end{equation}
and concatenates real and imaginary part for the subsequent feature extraction:
\begin{equation}
\text { ComplexToReal }: \mathbb{C}^{H \times \frac{W}{2} \times \frac{C}{2}} \rightarrow \mathbb{R}^{H \times \frac{W}{2} \times C},
\end{equation}
then we applies the convolution block with the kernel size of $1 \times 1$ in the frequency domain for capturing the long-distance dependence:
\begin{equation}
\operatorname{Conv} \circ \operatorname{ELU} \circ \operatorname{Conv}: \mathbb{R}^{H \times \frac{W}{2} \times  C} \rightarrow \mathbb{R}^{H \times \frac{W}{2} \times C}
\end{equation}
finally, utilize inverse transform to recover the spatial structure of features:
\begin{equation}
\begin{aligned}
\text { RealToComplex }: \mathbb{R}^{H \times \frac{W}{2} \times C} & \rightarrow \mathbb{C}^{H \times \frac{W}{2} \times \frac{C}{2}} \\
\text { Inverse Real FFT2d }: \mathbb{C}^{H \times \frac{W}{2} \times \frac{C}{2}} & \rightarrow \mathbb{R}_{ffc}^{H \times W \times \frac{C}{2}},
\end{aligned}
\end{equation}
And we utilize the multi-scale convolution blocks as local branch to capture the local information, as more important as global information in detail restoration procedure:
\begin{equation}
\begin{aligned}
\operatorname{Conv} \circ \operatorname{ELU} \circ \operatorname{Conv} with 1 \times 1: \mathbb{R}_{l}^{H \times W \times \frac{C}{2}} \rightarrow \mathbb{R}_3^{H \times W \times \frac{C}{2}}  \\
\operatorname{Conv} \circ \operatorname{ELU} \circ \operatorname{Conv} with 3 \times 3: \mathbb{R}_{l}^{H \times W \times \frac{C}{2}} \rightarrow \mathbb{R}_1^{H \times W \times \frac{C}{2}},
\end{aligned}
\end{equation}

Finally, we fusion all features from every branch and perform local residual learning by concatenate and add operator:
\begin{gather}
    \operatorname{Fusion}: \mathbb{R}_{f f c}^{H \times W \times \frac{C}{2}}+\mathbb{R}_{3}^{H \times W \times \frac{C}{2}}+\mathbb{R}_{1}^{H \times W \times \frac{C}{2}}
    + \mathbb{R}^{H \times W \times C} \\
    \rightarrow \mathbb{R}_{out}^{H \times W \times C},
\end{gather}

The presented MST block leverages the advantages of multi-scale convolution branches at local processing and FFC at global interaction, which achieves excellent trade-off in complexity and modeling-ability.

%The proposed MST block effectively captures the global context information and local features, which achieves excellent trade-off in complexity and modeling-ability due to the channel splitting solution and fast Fourier convolution.

\paragraph{\textbf{Cross-Layer Activation Gated Module (CLAGM).}}
The information loss caused by repetitive down-sampling seems unavoidable in a multi-scale feature flow network. Inspired by the recent research conclusion ~\cite{schiapparelli2022proteomic} that retinal ganglion cells and cerebral visual cortex cells have a whole new way of cell communication, which is mediated by exosomes different from synapses, we find that the additional external information can bring the valuable activation like an instructor for the current layer. Thus we hope that external information from the down-sampled original image to effectively regulate the input features from another layer. As shown in Fig.\ref{fig:overviewofarchitecture}(e),we perform a series of operations on $I_{in}^{\frac{1}{2}}$ or $I_{in}^{\frac{1}{4}}$ to generate the activation gated map $\mathcal{M}^{H\times W \times C}$:
\begin{equation}
    \mathcal{M}^{H\times W \times C} = \textbf{CLAGM}(I_{in}^{H\times W \times 3}),
\end{equation}
where the $\mathcal{M}^{H\times W \times C}$ is a unique attention map from additional information, whose size is the same as the $\mathbb{R}_{in}$. Then we multiply the $\mathcal{M}$ with input feature $\mathbb{R}_{in}$:
\begin{equation}
    \mathbb{R}_{gated}^{H\times W \times C} = \mathcal{M}^{H\times W \times C} \cdot \mathbb{R}_{in}^{H\times W \times C},
\end{equation}
then, the original $\mathbb{R}_{in}$ and weighted $\mathbb{R}_{in}^{'}$ are fusioned by channel merging and compressing:
\begin{equation}
\text { Conv with } 3 \times 3: \mathbb{R}_{in}^{H\times W \times C} + \mathbb{R}_{gated}^{H\times W \times C}  \rightarrow \mathbb{R}_{out}^{H\times W \times C}
\end{equation}

We found that this cross-scale mechanism can help the network better exploit degraded images' inner knowledge and optimize the information flow over the network.

\subsection{ Lower-level Layers}
First of all, the designed  Lower-level Layers as the level one role to extract the features from original degenerated images in original resolution. As shown in Fig. ~\ref{fig:overviewofarchitecture} (a), a $3 \times 3$ convolution is used to extract the features with $32$ channels ($16$ channels in the tiny version). Secondary,these features will be sent into Intermediate Layers and subsequent attention module. Then, we utilize the concatenate operation to fusion the features that feed back from Intermediate Layers and the refined features by attention module, thus aggregating the features with 64 channels (32 channels in the tiny version.):
\begin{equation}
    \operatorname{Concatenate:} \mathbb{R}_{S}^{H \times W \times 32} + \mathbb{R}_{I}^{H \times W \times 32} \rightarrow
    \mathbb{R}_{S}^{H \times W \times 64},
\end{equation}
where the $\mathbb{R}_{I}^{H \times W \times 32}$ is the features from Intermediate Layers. Then we perform the $3 \times 3$ convolution to compress the channel of features.
\begin{equation}
    \operatorname{Conv} with 3 \times 3:  \mathbb{R}_{S}^{H \times W \times 64} \rightarrow \mathbb{R}_{S}^{H \times W \times 32},
\end{equation}
As shown in Fig.\ref{fig:overviewofarchitecture} (a), two cascaded MST blocks and a $3\times 3$ convolution are used for generating the final clean image $J$ from features.

\subsection{Intermediate Layers}
As shown in Fig.\ref{fig:overviewofarchitecture} (a), The Intermediate Layers we structured as a transmitter and a receiver coordinates the complex information flow and enhances the coupling of each layer in the expert network. Specifically, the number of the channel of the MST blocks and Dual-Pool attention module only is $64$ channels ($32$ channels in tiny version) in this layer, and the CLAGM receives the $\frac{1}{2}$ down-sampled degenerated images to aggregate and refines features. In our expert network, the Intermediate Layers plays a vital role as a powerful connection between different layers, owing most complex features from different scales and refining them for exploring deep, valuable features.I
\vspace{-0.1cm}
\subsection{High-level Layers}
High-level Layers is responsible for exacting more information in high dimension space from features with $128$ channels ($64$ channels in tiny version) and $\frac{1}{4}$ size. As shown in Fig.\ref{fig:overviewofarchitecture} (a), with the help of $\frac{1}{4}$ down-sampled degenerated images and external features from Intermediate Layers, High-level Layers helps the designed expert network to generate more pleasant clean images from degraded winter images.
\begin{table*}[t!]
\caption{\small{Quantitative comparisons of our expert networks and DAN-Net(Tiny) with the state-of-the-art desnowing methods on CSD,SRRS and Snow 100K desnowing dataset (PSNR(dB)/SSIM). The best results are shown in \textbf{bold}, and second best results are \underline{underlined}.}}
		\centering
		 \resizebox{10cm}{!}{
\begin{tabular}{l|cc|cc|cc|c|c}
\toprule[1.2pt]

\multicolumn{1}{c|}{\multirow{2}{*}{Method}} & \multicolumn{2}{c|}{CSD(2000)} & \multicolumn{2}{c|}{SRRS (2000)} & \multicolumn{2}{c|}{Snow 100K (2000)} & \multirow{2}{*}{\#Param} & \multirow{2}{*}{\#GMacs}\\ \cline{2-7}
\multicolumn{1}{c|}{}                        & PSNR           & SSIM          & PSNR            & SSIM           & PSNR               & SSIM             &                          \\ \hline
(TIP'18)Desnow-Net~\cite{liu2018desnownet}                                   & 20.13          & 0.81         & 20.38           & 0.84           & 30.50              & 0.94             & 15.6M           &-           \\
(ICCV'17)CycleGAN~\cite{engin2018cycle}                                     & 20.98          & 0.80          & 20.21           & 0.74           & 26.81              & 0.89             & 7.84M        &42.38G     \\

(CVPR'20)All-in-One~\cite{li2020all}                                   & 26.31          & 0.87          & 24.98           & 0.88           & 26.07              & 0.88             & 44 M             &12.26G           \\
(ECCV'20)JSTASR~\cite{chen2020jstasr}                                       & 27.96          & 0.88          & 25.82           & 0.89           & 23.12              & 0.86             & 65M           &-         \\
(ICCV'21)HDCW-Net~\cite{chen2021all}                                     & \underline{29.06}    & \underline{0.91}    & \underline{27.78}     & \underline{0.92}     & \underline{31.54}        &\underline{0.95}       & 699k     &9.78G   \\ \hline
\gr Desnowing Expert Net                               & 30.56          &\textbf{0.95}         & 29.07             & \textbf{0.95}        &32.14                  &\textbf{0.96}                & 1.1M        & 12.00G                \\
\gr Desnowing Expert Net-Tiny                              & {29.06}          & {0.92}          & {28.20}               &{0.94}              & {31.67}                  & {0.95}                & 288K    & 3.06G                 \\ 
\gr DAN-Net                              & \textbf{30.82}          & \textbf{0.95}          & \textbf{29.34}               & \textbf{0.95}          & \textbf{32.48}                  &\textbf{0.96}                & 2.73M        & 31.36G                \\
\gr DAN-Net-Tiny                              & {29.12}          & {0.92}          & {28.32}               &{0.94}              & {31.93}                  &{0.95}                & 1.02M    & 13.47G                 \\ 
\bottomrule
\end{tabular}
	}
\label{tab:ComparisonDesnowing}
			
\end{table*}

\section{Degradation-Adaptive Neural Network}
\begin{figure*}[t]
% \vspace{-1mm}
% \setlength{\abovecaptionskip}{0.0cm} %调整caption与图的距离
% \setlength{\belowcaptionskip}{-0.2cm}   %调整图片标题与下文距离
    \centering
    \includegraphics[width=14cm]{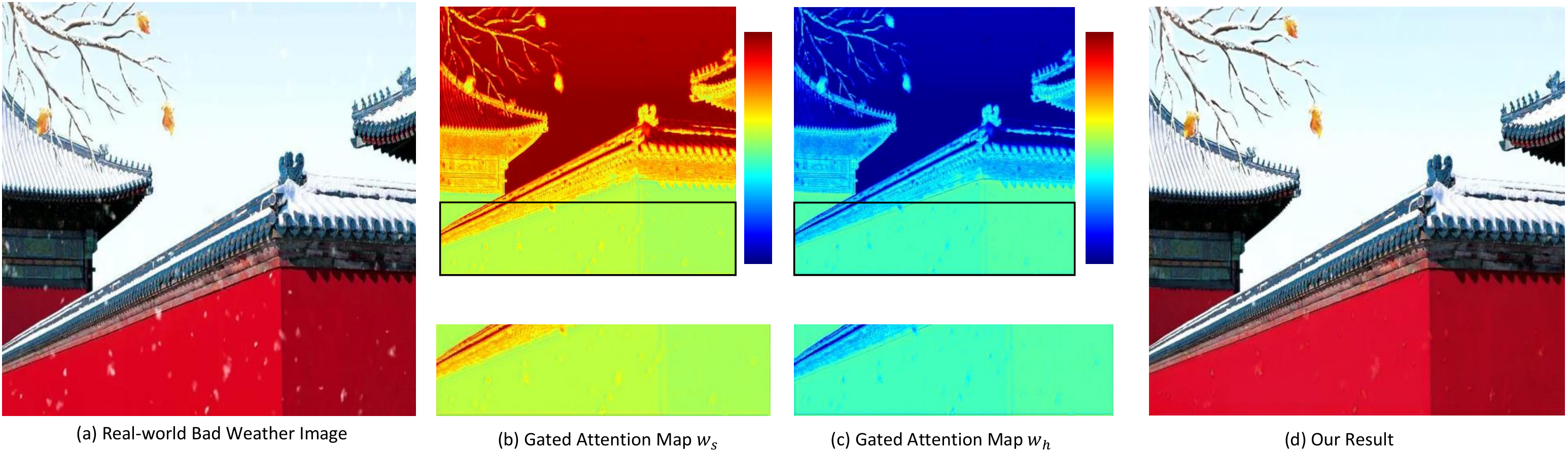}
    \caption{\small{Illustration of the real-world degenerated image, two Gated Attention Maps $w_s$, $w_h$ corresponding to the desnowing and dehazing expert activation intensity, respectively, and the restoration result of our DAN-Net. Maps are visualized by ColorJet for observation and comparison. Please note the local detail comparison of gated attention maps.}}
    \label{fig:gatedattentionmap}
\end{figure*}

The mixture of experts (MoE) strategy~\cite{jacobs1991adaptive,jordan1994hierarchical,gross2017hard,he2021fastmoe,pavlitskaya2020using} is a long-standing solution that utilizes the integration of multiple expert networks to improve single-task performance. A trainable gating network is usually employed to produce the gated weights for activating each expert network based on an explicit or implicit partition of the input data, for instance, the labels or classes (explicit) or contextual content features (implicit). In the proposed DAN-Net, the Adaptive Gated Neural learns how to generate the gated attention maps as the activating weight, according to the uncertain degradation of the input winter image without explicit signal. Besides, instead of the intent of improving single-task performance in previous methods~\cite{pavlitskaya2020using,aljundi2017expert}, we utilize the adaptive mechanism based on MoE and the rich inner knowledge of task-specific experts not only to address the uncertain multi-degradation problem but also to boost the comprehensive performance of DAN-Net in winter scenes. Such a novel image restoration pipeline is effective and efficient due to the degradation-adaptive mechanism and well-learned expert networks with lightweight architecture. 

Specifically, we design Adaptive Gated Neural as the guider to adaptively control the contribution of two different pre-trained expert networks. We observe the uneven spatial distribution and different types of degradations in a single degraded image of real-world natural winter conditions, thus utilizing the 2D attention map as the gated map to address these degraded attributes for better performance. As shown in Fig.~\ref{fig:overviewofarchitecture} (b) and (f), the Adaptive Gated Neural predicts gated attention map $w_{s} \in \mathbb{R}^{1 \times H \times W},w_{h} \in \mathbb{R}^{1 \times H \times W}$ based on original degenerated images:
\begin{equation}
    w_h,w_s = \textbf{AGN}(I_{in}),
\end{equation}
where the $\textbf{AGN}$ denotes the Adaptive Gated Neural, and the $w_h,w_s$ denote the gated attention map for dehazing and desnowing results, respectively. Then we multiply the gated attention map $w_h$ and $w_s$ with $J_{dehazing},J_{desnowing}$ generated by dehazing expert and desnowing expert networks, respectively:

\begin{equation}
    J_{out} = w_h \cdot J_{dehazing} + w_s\cdot J_{desnowing} 
\end{equation}
where the $J_{out}^{3 \times H \times W}$ is the final result of the proposed DAN-Net. Our experiments demonstrate that the adaptive gated mechanism based on attention maps we proposed is beneficial for the performance of both dehazing and desnowing tasks, our DAN-Net achieves better performance on both tasks than single task-specific expert network. Moreover, we illustrate the gated attention maps of the real-world snowy image.  As shown in Fig.~\ref{fig:gatedattentionmap}, the snowy degradation regions in $w_s$ have higher activation intensity than $w_h$ and other areas.
\section{Loss Function}
For further exploring the performance of our method, we utilize Spectral Transform loss function and $L_1$ loss function as the combination for the network training.
\subsection{Basic Reconstruction Loss}
We only use Charbonnier loss ~\cite{charbonnier1994two} as our basic reconstruction loss:
\begin{equation}
\mathcal{L}_{\text {char }}=\frac{1}{N} \sum_{i=1}^{N} \sqrt{\left\|X^{i}-Y^{i}\right\|^{2}+\epsilon^{2}},
\end{equation}
with constant $\epsilon$ emiprically set to $1e^-3$ for all experiments, $X^i$ denotes the generated image of network and $Y^i$ correspondingly denotes the ground truth.
\subsection{Spectral Transform Loss}
We utilize the spectral transform to evaluate the difference of the reconstructed images and the ground truth images in the frequency domain:
\begin{equation}
\mathcal{L}_{st} = \left\|\mathcal{F} \mathcal{T}\left( \mathbf{J_{clean}}\right)-\mathcal{F} \mathcal{T}(\mathbf {J_{gt}})\right\|_{1},
\end{equation}
where $\mathcal{F}\mathcal{T}$ denotes the FFT operation, $\mathbf{J_{clean}}$ denotes the generated images of our network, and $\mathbf{J_{gt}}$ denotes the ground truth images. We found that the spectral transform loss function is markedly beneficial to the improvement of SSIM. Please refer to our experiment section for the detail ablation study about $\mathcal{L}_{st}$ function.
\subsection{Overall Loss Function}
The overall loss function of networks are defined as follow:
\begin{equation}
    \mathcal{L}_{expert} = \mathcal{L}_{char} + \lambda_{st} \mathcal{L}_{st}
\end{equation}
where $\lambda_{st}$ is the trade-off weight. 
\section{Experiment}
\begin{figure}[h]
    \centering
    \includegraphics[width=8.0cm]{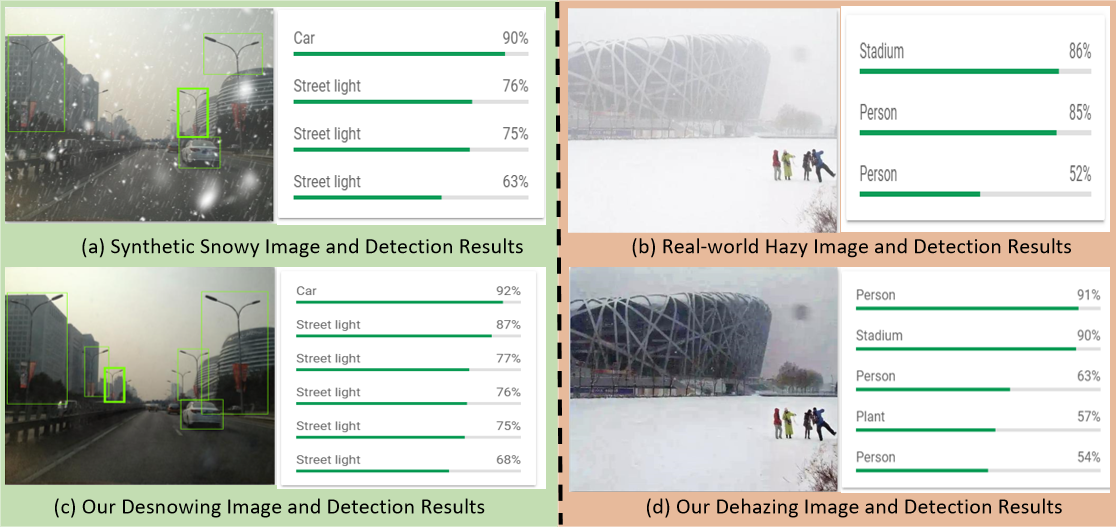}
    \caption{\small{Synthetic and Realistic winter degenerated images (top) and corresponding restoration results (bottom) of the proposed DAN-Net with labels and confidences supported by Google Vision API. The real-world hazy image (b) is from the RWSD we collected.}}
    \label{fig:GoogleCVAPI}
\end{figure}

\begin{table*}[t]
\caption{\small{Quantitative comparisons of our framework with the state-of-the-art single image dehazing methods on Haze4k and SOTS outdoor dataset (PSNR(dB)/SSIM). The best results are \textbf{bold}, second best results are \underline{underlined}.}}
		\centering
		 \resizebox{9cm}{!}{
		\begin{tabular}{l|cc|cc|c|c}
		\toprule[1.2pt]
			\multirow{2}{*}{Method}             &      \multicolumn{2}{c|}{Haze4k~\cite{chen2021all}}  &       \multicolumn{2}{c|}{SOTS Outdoor}   & \multirow{2}{*}{\#Param} &
			\multirow{2}{*}{\#GMacs}  \\ \cline{2-5}
			
		& {PSNR$\uparrow$}  & {SSIM$\uparrow$}  & {PSNR$\uparrow$}  & {SSIM$\uparrow$} \\[1pt] \hline \hline
			(TPAMI'10)DCP ~\cite{he2010single}  &       14.01       &       0.76       &         15.09         & 0.76 &- &-  \\[1pt]
			
			(CVPR'16)NLD~\cite{berman2016non}     &15.27 &0.67 &17.27 &0.75 & - & -\\[1pt]
			
			(TIP'16)DehazeNet ~\cite{cai2016dehazenet}     &       19.12       &       0.84       &       22.46       &       0.85   & 0.01M & 0.52G \\[1pt]
			(ICCV'17)GDN ~\cite{griddehazenet}     &         23.29         &         0.93         &       30.86       &       0.98      & 0.96M &21.49G\\[1pt]
			(CVPR'20)MSBDN ~\cite{msbdn}       &       22.99       &       0.85       &       23.36       &       0.87  & 31.35M & 41.58G \\[1pt]
			(CVPR'20)DA       &       24.03       &       0.90       &       27.76       &0.938 & 1.64M  &33.59G\\[1pt]
			(AAAI'20)FFA-Net ~\cite{ffa-net}                  &       26.97       &       0.95       &  \underline{33.57}       &\underline{0.98}  &4.6M&150.94G \\[1pt]
			(CVPR'21 Oral)PSD-FFA~\cite{chen2021psd}                  & 16.56  &0.73       & 15.29       & 0.72  &4.6M&150.94G \\[1pt]
			
			(ACMMM'21)DMT-Net ~\cite{liu2021synthetic}                   &\underline{28.53}       & \underline{0.96} &- &-   & 54.9 M & 80.71G\\[1pt] 
		 \hline
\gr	 Dehazing Expert-Net  &{29.12} &\textbf{0.97} &{33.59}       &\textbf{0.99}  &1.14M& 12.00G \\[1pt] 
\gr	 Dehazing Expert-Net-Tiny   &{28.18} &{0.96} &{32.09} &{0.98}  &288K& 3.06G\\[1pt] 
\gr	 DAN-Net  &\textbf{29.24} &\textbf{0.97} &\textbf{33.69}       &\textbf{0.99} &2.73M& 31.36G \\[1pt] 
\gr      DAN-Net-Tiny &{28.56} &{0.96} &{32.24}       &{0.98}  &1.02M& 13.47G \\[1pt] 
     
	 %DAN-Net &\textbf{29.24} &\textbf{0.97} 
	 \bottomrule
		\end{tabular}
	}
\label{tab:ComparisonDehazing}
\end{table*}

%\footnotetext[1]{Google Vision API:\url{https://cloud.google.com/vision/}}

\subsection{Datasets and Metrics}
We choose the widely used PSNR and SSIM as experimental metrcs to measure the performance of our networks. We train and test our expert networks on five large datasets : RESIDE~~\cite{SOTS}, Haze4k~\cite{liu2021synthetic}, CSD~\cite{chen2021all}, SRRS~\cite{chen2020jstasr} and Snow100K~\cite{liu2018desnownet}, following the benchmark-setting of latest dehazing or desnowing methods~\cite{liu2021synthetic,chen2021all} for authoritative evaluation. And we randomly choose 1,000 paired data from hazy or snowy datasets respectively to train our DAN-Net.

\subsection{Traning Settings}
We augment the training dataset with randomly rotated by 90, 180, 270 degrees and horizontal flip. The training image patches with the size $256 \times 256$ are extracted as input images of our networks. We utilize Adam optimizer with initial learning rate of $2 \times 10^{-4}$, and adopt the CyclicLR to adjust the learning rate, where on the triangular mode, the value of gamma is 1.0, base momentum is 0.9, max learning rate is $3 \times 10^{-4}$ and base learning rate is the same as initial learning rate. We use PyTorch~\cite{automatic} to implement our all networks with 4 Tesla V100 GPU with total batchsize of 40.
The $\lambda_{st}$ as trade-off weight is set as 0.2 in all experiments. 

\vspace{-1em}
\subsection{Comparison with State-of-the-art Methods}
~\paragraph{Dehazing Results.} In Table~\ref{fig:GoogleCVAPI}, we summarize the performance of our Dehazing expert network (Tiny), DAN-Net (Tiny) and SOTA methods on Haze4k ~\cite{liu2021synthetic} and RESIDE dataset ~\cite{SOTS} (\emph{a.k.a}, SOTS). Our DAN-Net achieves the best performance with 29.24dB PSNR and 0.97 SSIM on the test dataset of Haze4k, cimpared to SOTA methods. The tiny version of DAN-Net also achieves 28.62 dB PSNR and 0.96 SSIM on the test dataset of Haze4k with impressive fewer computation and parameters.

In particular, compare to the DMT-Net with the top performance of previous methods, our DAN-Net achieves 0.89dB PSNR performance gains with the significant reduction of 52.6M parameters.

\begin{figure*}[!h]
% \vspace{-1mm}
% \setlength{\abovecaptionskip}{-0.1cm} %调整caption与图的距离
% \setlength{\belowcaptionskip}{-0.4cm}   %调整图片标题与下文距离
    \begin{center}
        \begin{tabular}{ccccccccccc}
\includegraphics[width = 0.09\linewidth]{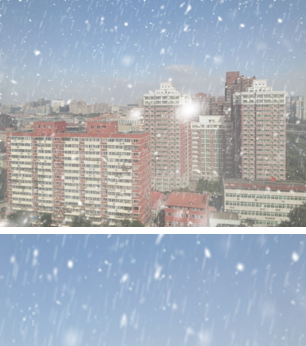} &\hspace{-2.5mm}
\includegraphics[width = 0.09\linewidth]{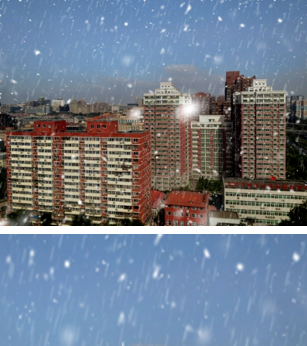} &\hspace{-2.5mm}
\includegraphics[width = 0.09\linewidth]{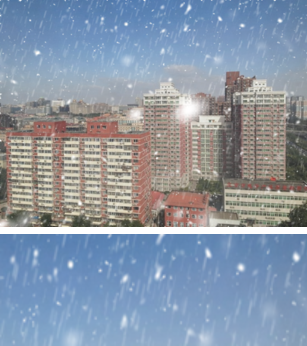} &\hspace{-2.5mm}
\includegraphics[width = 0.09\linewidth]{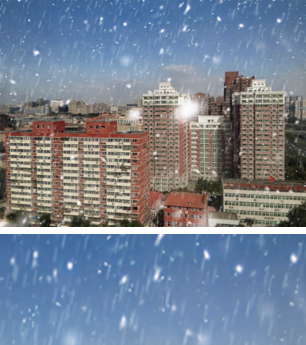} &\hspace{-2.5mm}

\includegraphics[width = 0.09\linewidth]{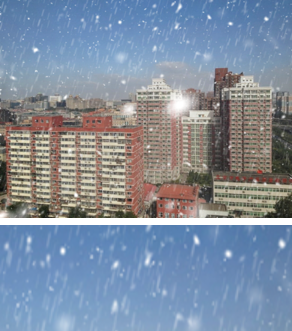} &\hspace{-2.5mm}
\includegraphics[width = 0.09\linewidth]{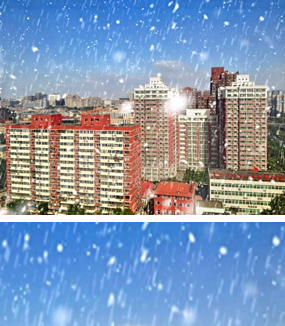} &\hspace{-2.5mm}
\includegraphics[width = 0.09\linewidth]{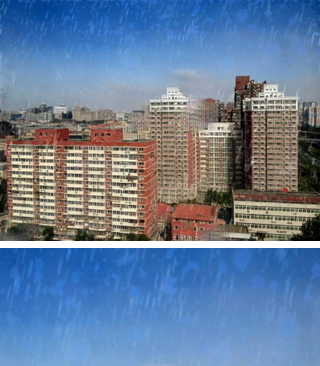} &\hspace{-2.5mm}
\includegraphics[width = 0.09\linewidth]{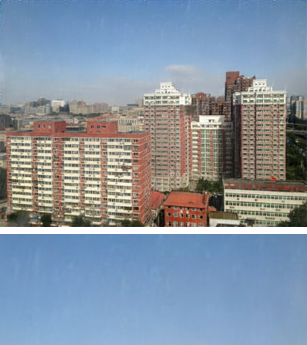} &\hspace{-2.5mm}
\includegraphics[width = 0.09\linewidth]{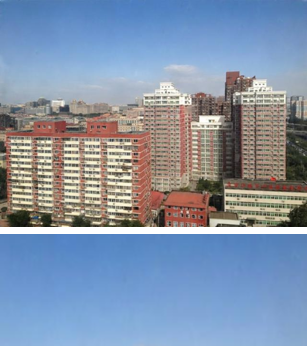} &\hspace{-2.5mm}
\includegraphics[width = 0.09\linewidth]{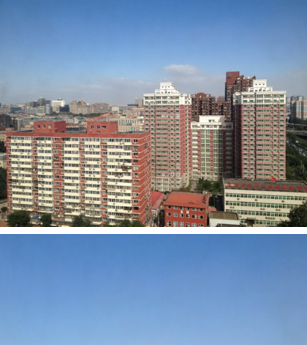} &\hspace{-2.5mm}
\vspace{1mm}\\
\includegraphics[width = 0.09\linewidth]{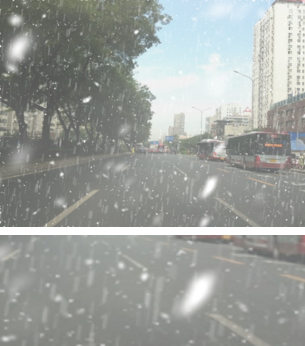} &\hspace{-2.5mm}
\includegraphics[width = 0.09\linewidth]{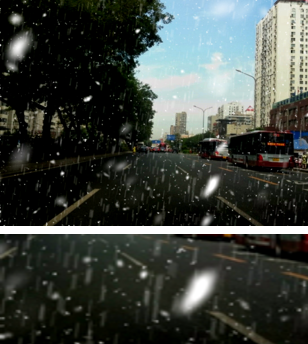} &\hspace{-2.5mm}
\includegraphics[width = 0.09\linewidth]{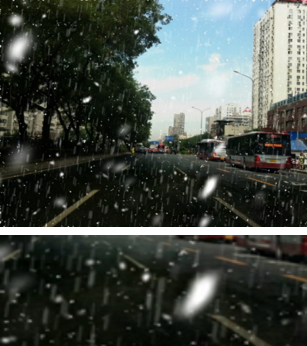} &\hspace{-2.5mm}
\includegraphics[width = 0.09\linewidth]{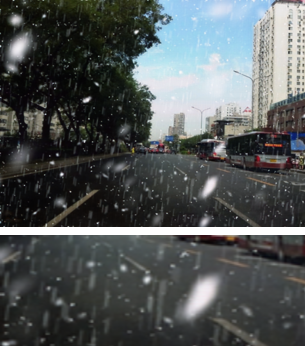} &\hspace{-2.5mm}

\includegraphics[width = 0.09\linewidth]{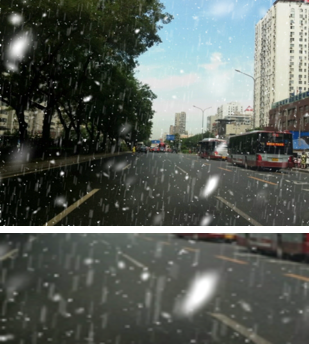} &\hspace{-2.5mm}
\includegraphics[width = 0.09\linewidth]{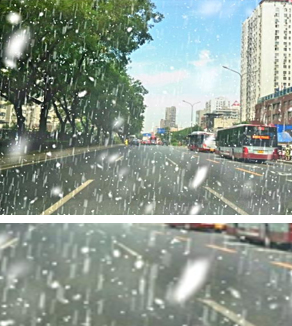} &\hspace{-2.5mm}
\includegraphics[width = 0.09\linewidth]{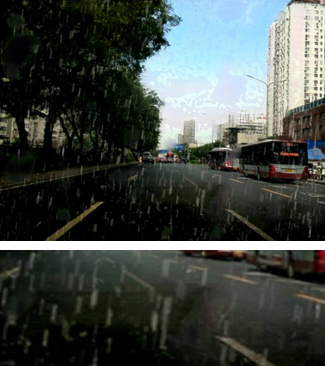} &\hspace{-2.5mm}
\includegraphics[width = 0.09\linewidth]{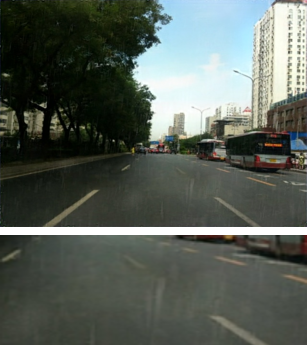} &\hspace{-2.5mm}
\includegraphics[width = 0.09\linewidth]{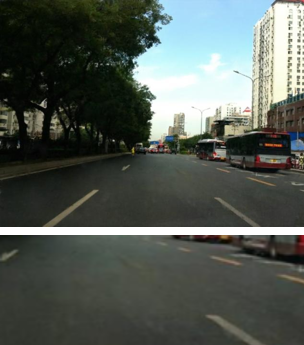} &\hspace{-2.5mm}
\includegraphics[width = 0.09\linewidth]{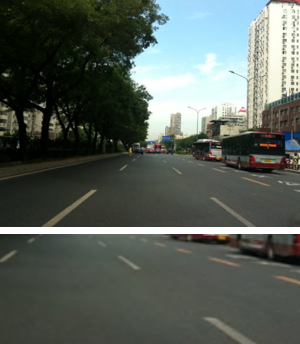}&\hspace{-2.5mm}
\vspace{1mm}\\
\includegraphics[width = 0.09\linewidth]{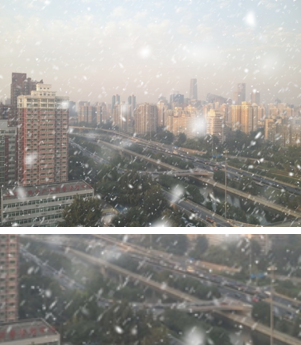} &\hspace{-2.5mm}
\includegraphics[width = 0.09\linewidth]{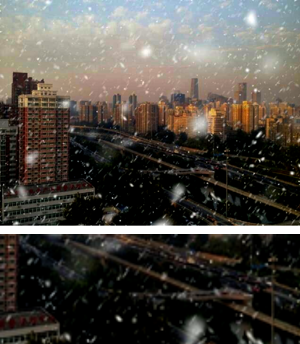} &\hspace{-2.5mm}
\includegraphics[width = 0.09\linewidth]{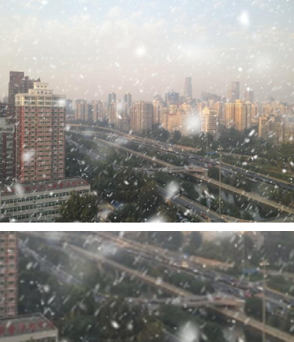} &\hspace{-2.5mm}
\includegraphics[width = 0.09\linewidth]{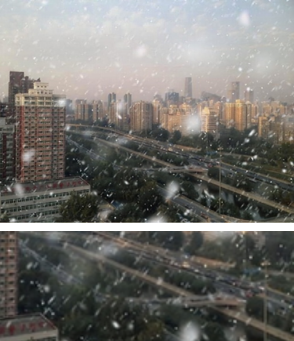} &\hspace{-2.5mm}
\includegraphics[width = 0.09\linewidth]{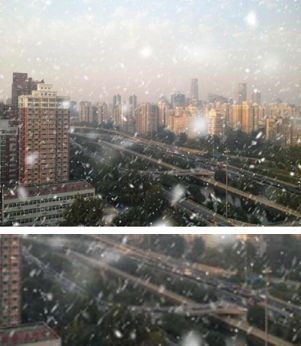} &\hspace{-2.5mm}
\includegraphics[width = 0.09\linewidth]{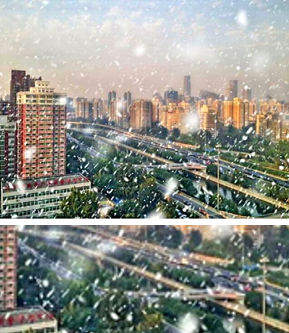} &\hspace{-2.5mm}
\includegraphics[width = 0.09\linewidth]{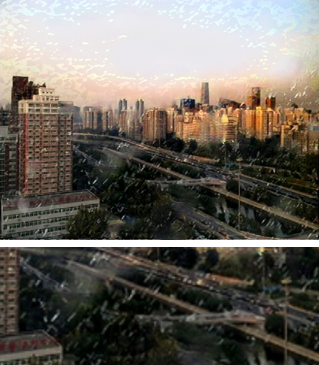} &\hspace{-2.5mm}
\includegraphics[width = 0.09\linewidth]{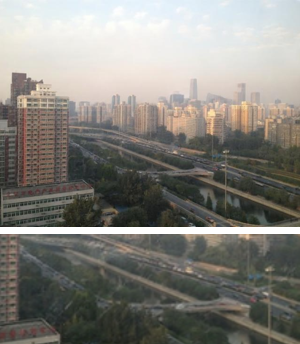} &\hspace{-2.5mm}
\includegraphics[width = 0.09\linewidth]{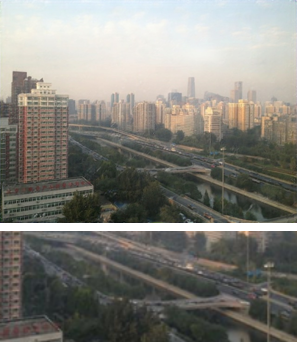} &\hspace{-2.5mm}
\includegraphics[width = 0.09\linewidth]{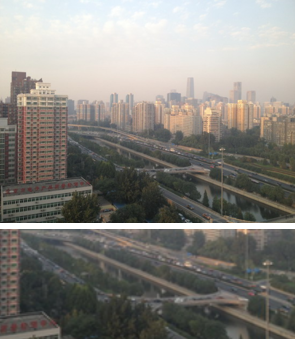}& \hspace{-2.5mm}
\\

\small{(a) Input}  &\hspace{-4mm} \small{(b) DehazeNet~\cite{cai2016dehazenet}}&\hspace{-4mm} \small{(c) FFA~\cite{ffa-net}} &\hspace{-4mm} \small{(d) DA~\cite{shao2020domain}} &\hspace{-4mm} (e)\small{MSBDN~\cite{msbdn}}&\hspace{-4mm} (f)\small{PSD-FFA~\cite{chen2021psd}}&\hspace{-4mm} (g)\small{JSTASR~\cite{chen2020jstasr}}&\hspace{-4mm} \small{(h)HDCW~\cite{chen2021all}} &\hspace{-4mm} \small{(i) DAN(Ours)}&\hspace{-4mm} \small{(j) Ground-truth} 
\\
\end{tabular}
\end{center}
\caption{\small{Visual comparisons on results of various methods (b-e) and our DAN-Net(f) on synthetic winter photos. For illustrating the limitations of previous image dehazing methods in winter scenes, we compare our method with the latest dehazing methods. Please zoom in for a better illustration.}}\label{fig:visualcomparison1}
\end{figure*}

~\paragraph{Desnowing Results.} We also summarize the performance of our methods and other desnowing methods on three widely used benchmarks in Table~\ref{tab:ComparisonDesnowing}. We compare the performance with previous state-of-the-art methods like HDCW-Net~\cite{chen2021all}. It's worth noting that we also compare the performance of our approach with All-in-One network~\cite{li2020all} which is trained to perform all the above tasks with a single model instance. Our method DAN-Net and its tiny version are also trained to perform both tasks using a single model instance.

\begin{figure*}[h!]
%     \vspace{-1mm}
%     \setlength{\abovecaptionskip}{-0.5cm} %调整caption与图的距离
% \setlength{\belowcaptionskip}{-0.8cm}   %调整图片标题与下文距离
    \begin{center}
        \begin{tabular}{cccccccccc}

\includegraphics[width = 0.1\linewidth]{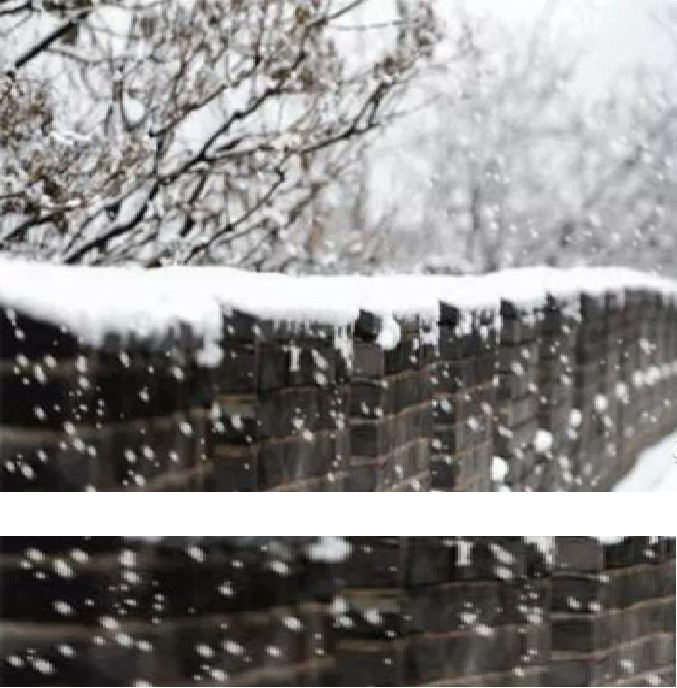} &\hspace{-2.5mm}
\includegraphics[width = 0.1\linewidth]{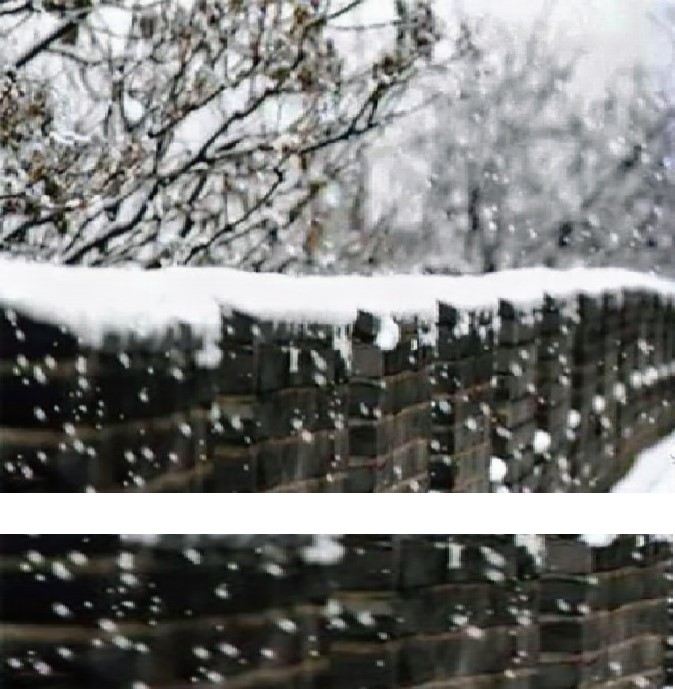} &\hspace{-2.5mm}
\includegraphics[width = 0.1\linewidth]{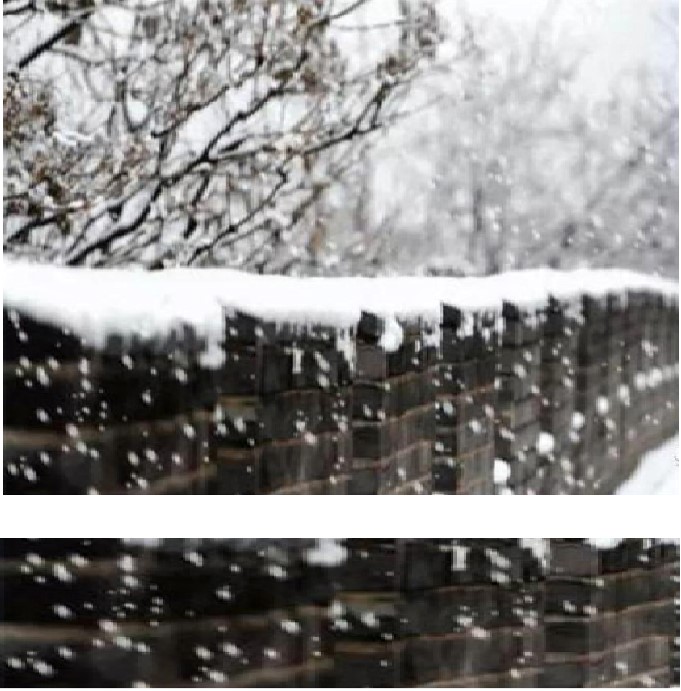} &\hspace{-2.5mm}
\includegraphics[width = 0.1\linewidth]{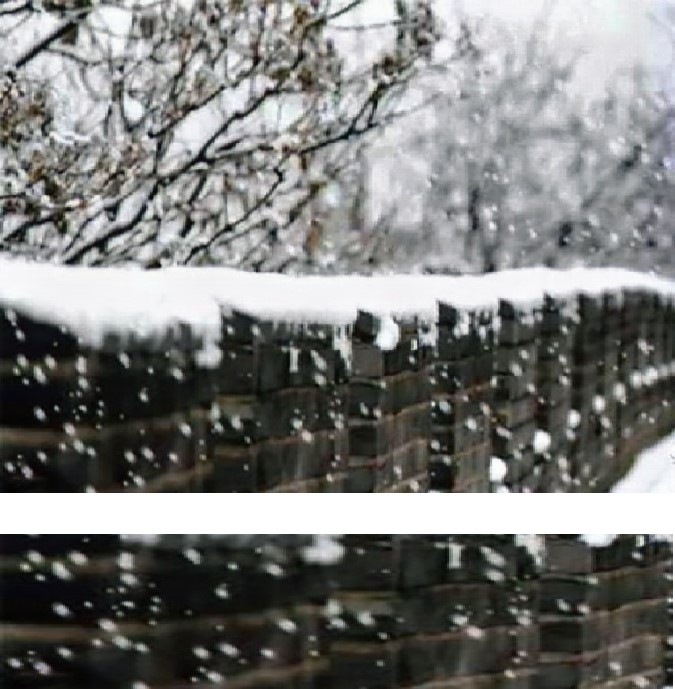} &\hspace{-2.5mm}

\includegraphics[width = 0.1\linewidth,height=0.1\linewidth]{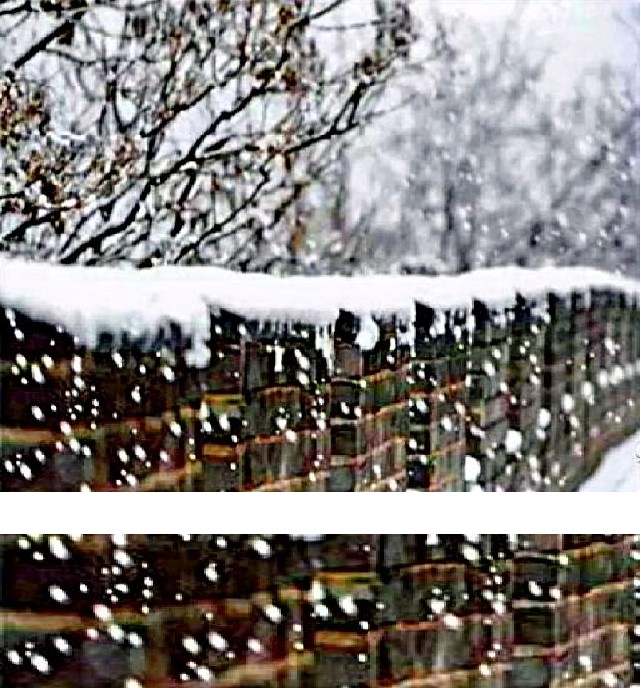} &\hspace{-2.5mm}

\includegraphics[width = 0.1\linewidth]{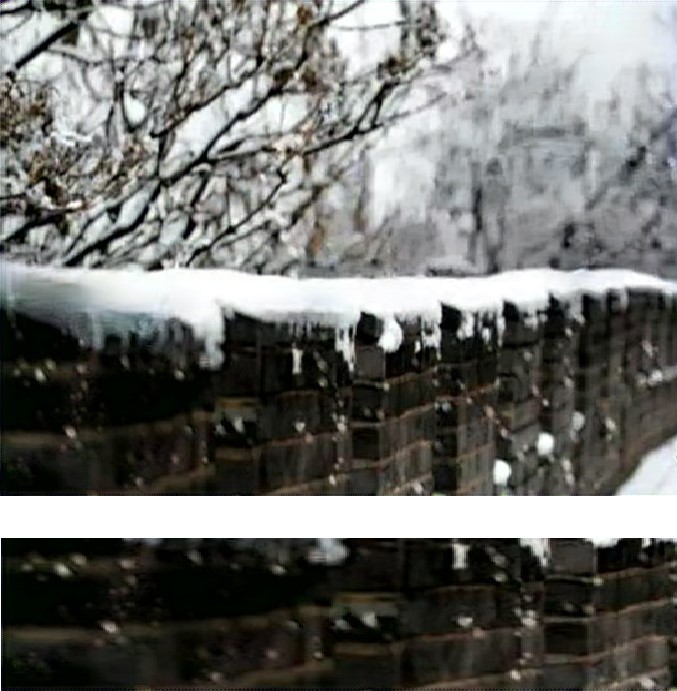} &\hspace{-2.5mm}
\includegraphics[width = 0.1\linewidth]{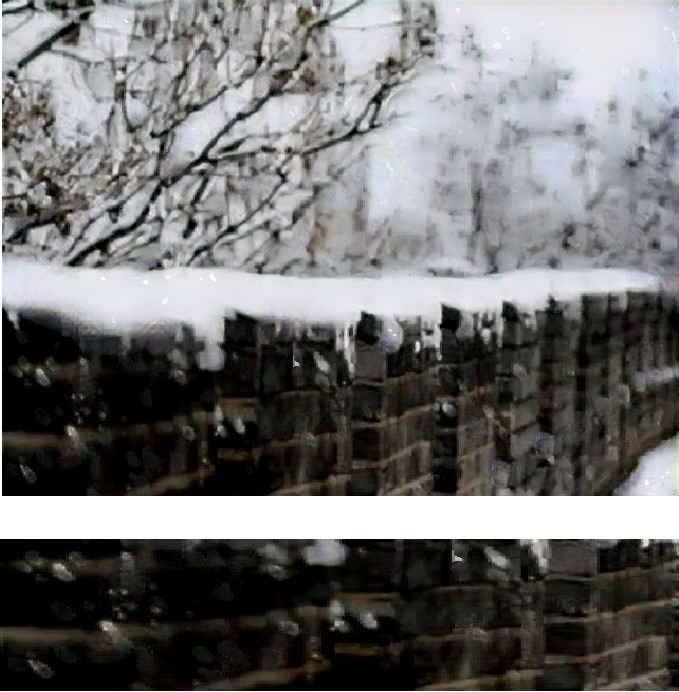} &\hspace{-2.5mm}
\includegraphics[width = 0.1\linewidth]{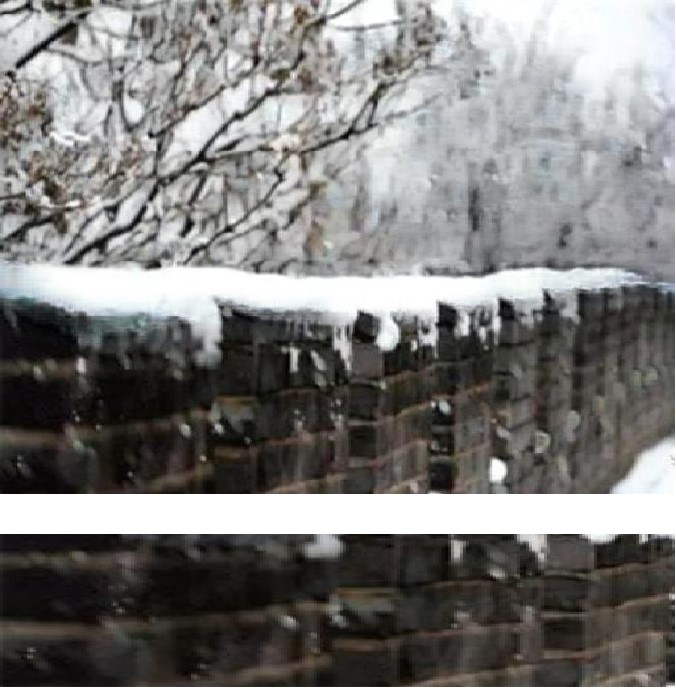} &\hspace{-2.5mm}
\includegraphics[width = 0.1\linewidth]{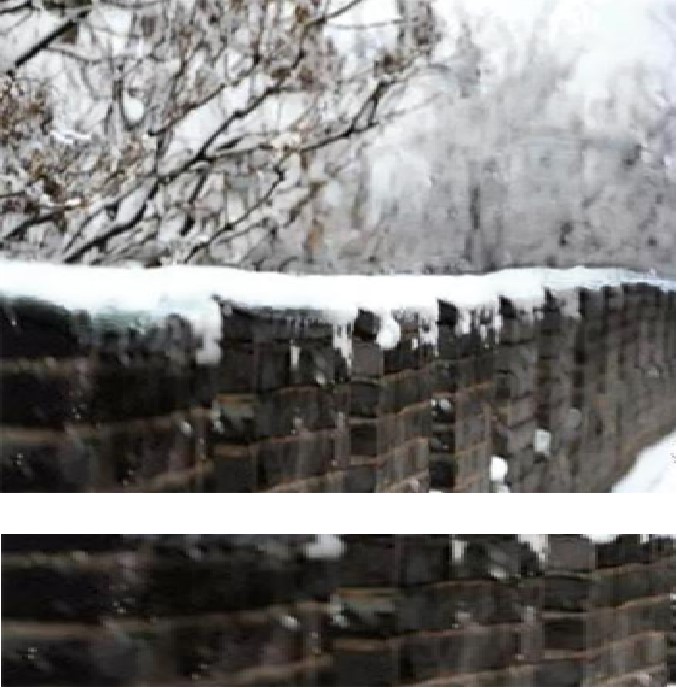} &\hspace{-2.5mm}

\vspace{1mm}\\

\includegraphics[width = 0.1\linewidth]{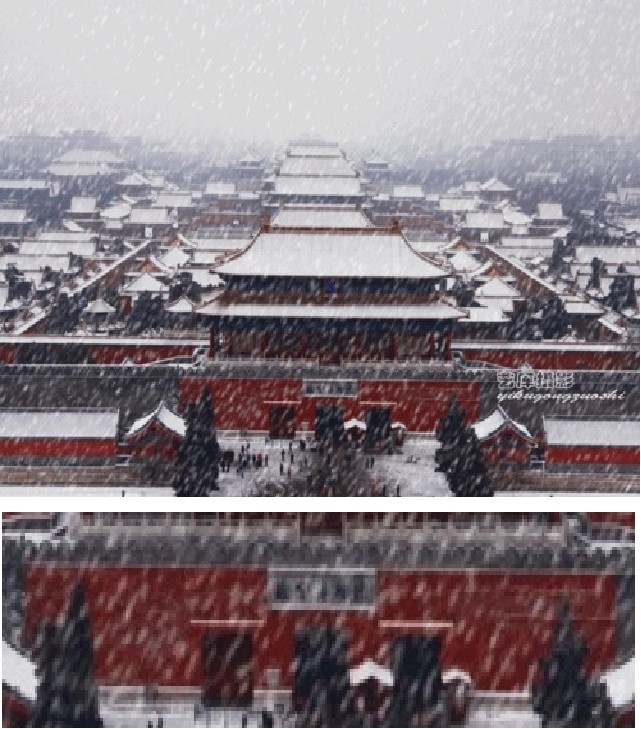} &\hspace{-2.5mm}
\includegraphics[width = 0.1\linewidth]{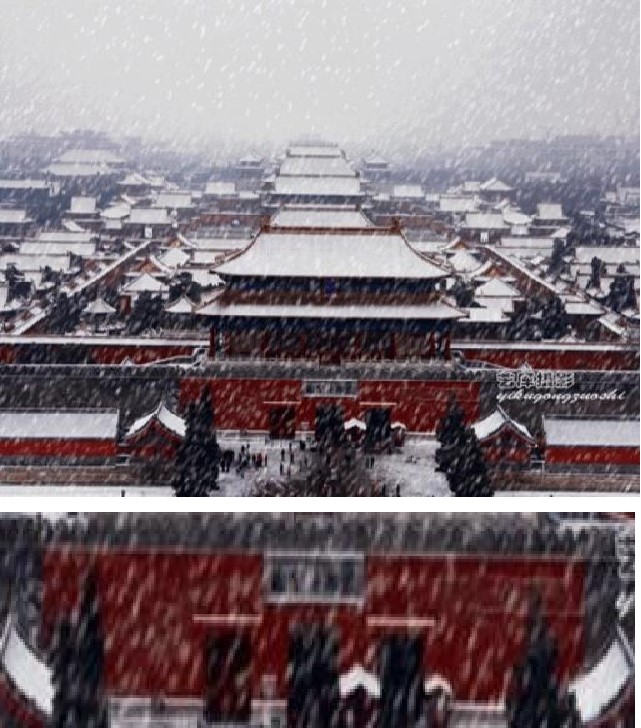} &\hspace{-2.5mm}
\includegraphics[width = 0.1\linewidth]{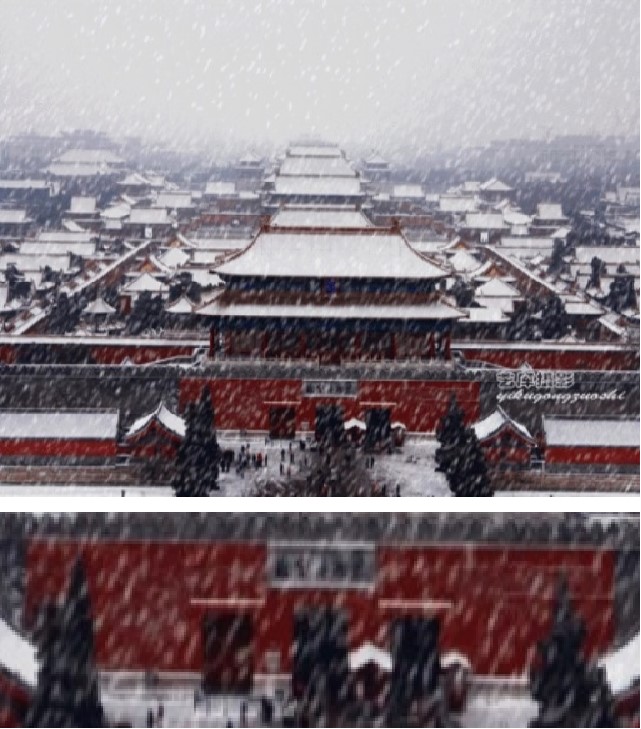}&\hspace{-2.5mm}
\includegraphics[width = 0.1\linewidth]{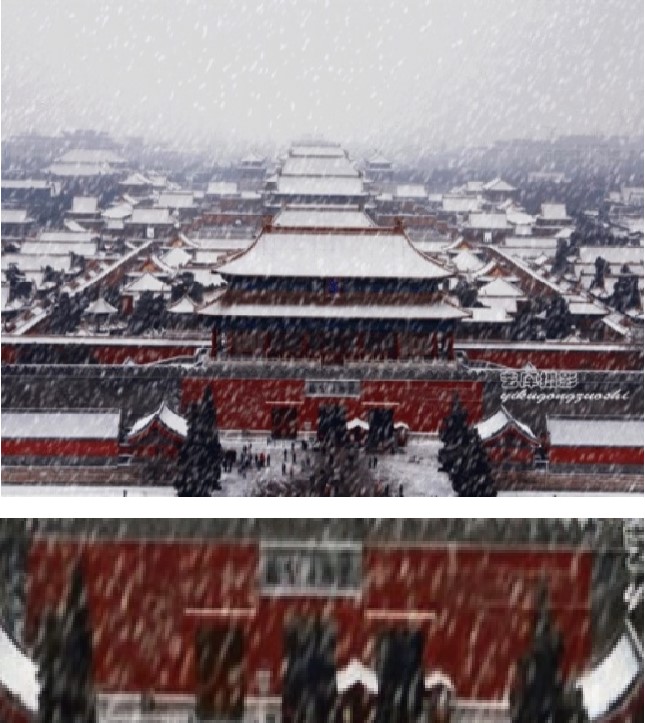} &\hspace{-2.5mm}

\includegraphics[width = 0.1\linewidth]{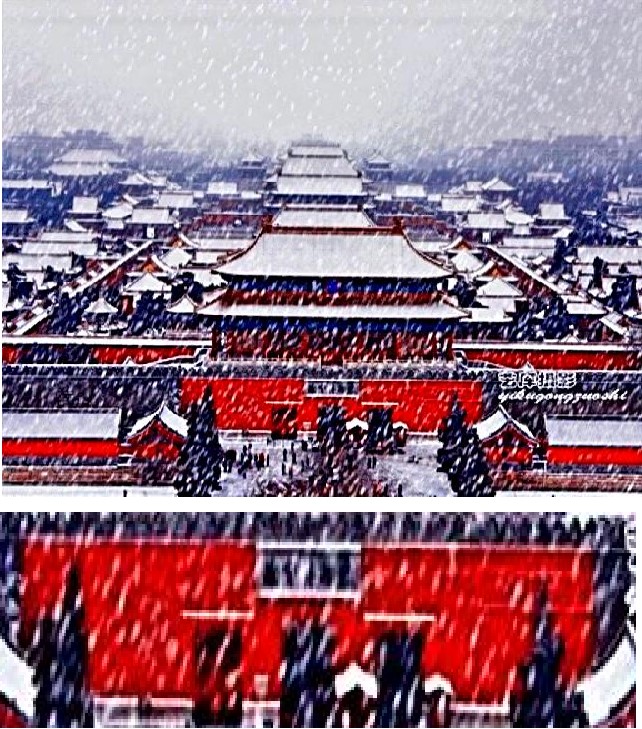} &\hspace{-2.5mm}

\includegraphics[width = 0.1\linewidth]{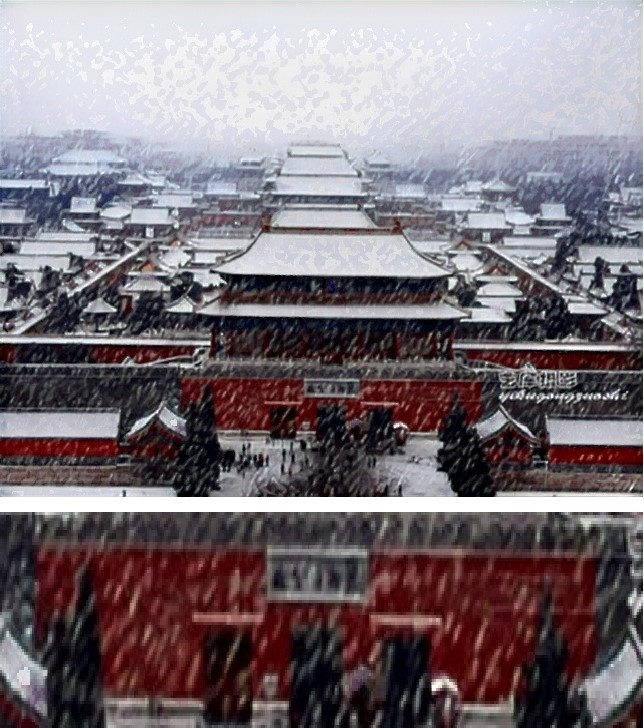} &\hspace{-2.5mm}
\includegraphics[width = 0.1\linewidth]{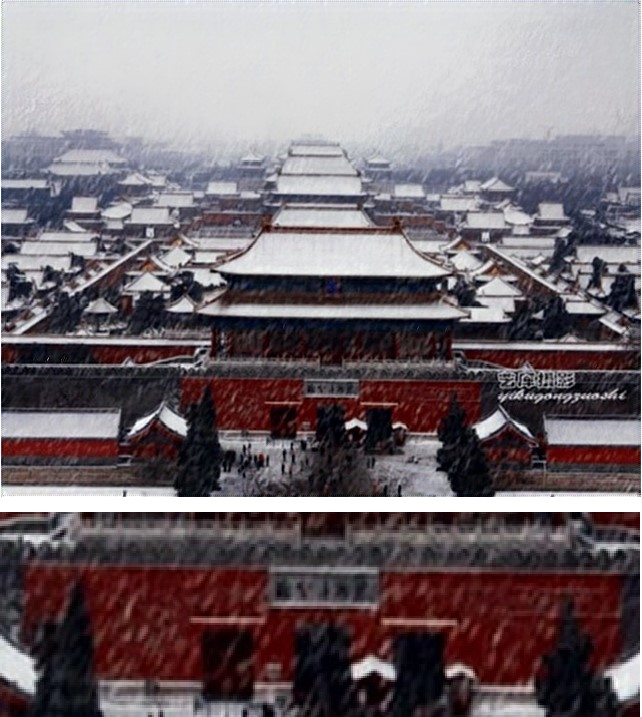} &\hspace{-2.5mm}
\includegraphics[width = 0.1\linewidth]{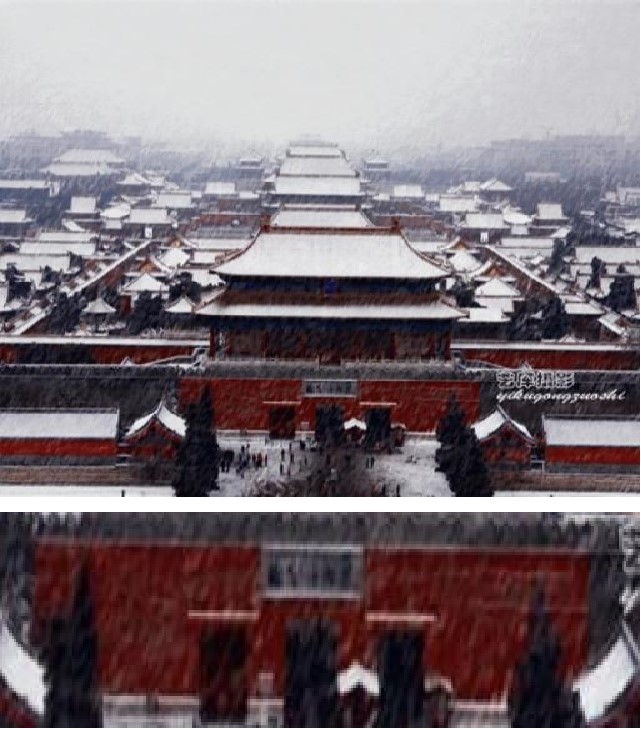} &\hspace{-2.5mm}

\includegraphics[width = 0.1\linewidth]{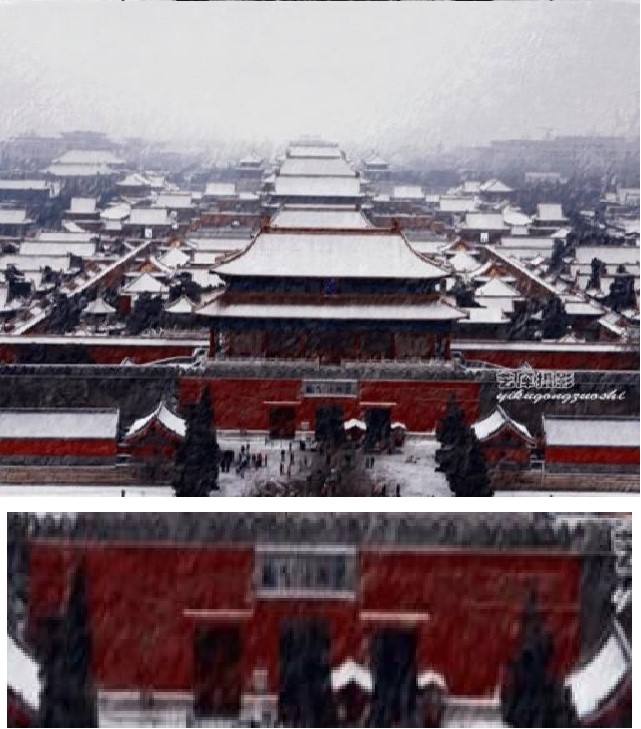} &\hspace{-2.5mm}

\\

\small{(a)Input}  &\hspace{-4mm} 
\small{(b)MSBDN~\cite{msbdn}} &\hspace{-4mm} 
\small{(c) FFA~\cite{ffa-net}} &\hspace{-4mm}  
\small{(d)DA~\cite{shao2020domain}} &\hspace{-4mm}
\small{(e)PSD-FFA~\cite{chen2021psd}} &\hspace{-4mm}
\small{(f)JSTASR~\cite{chen2020jstasr}} &\hspace{-4mm}
\small{(g)HDCW~\cite{chen2021all}} &\hspace{-4mm} 
\small{(h)DAN-Tiny(Ours)} &\hspace{-4mm} 
\small{(i)DAN(Ours)} &\hspace{-4mm} 
\\
        \end{tabular}
    \end{center}
\caption{\small{Visual comparisons on results produced by our DAN-Net(i), DAN-Net-Tiny(h) and SOTA image restoration methods(b-g) on real-world winter photos(a).}}\label{fig:visualcomparison2}
\end{figure*}

\begin{figure*}[t!]
%     \vspace{5mm}
%     \setlength{\abovecaptionskip}{-0.4cm} %调整caption与图的距离
% \setlength{\belowcaptionskip}{-0.4cm}   %调整图片标题与下文距离
    \begin{center}
        \begin{tabular}{ccccccccccc}

\includegraphics[width = 0.1\linewidth]{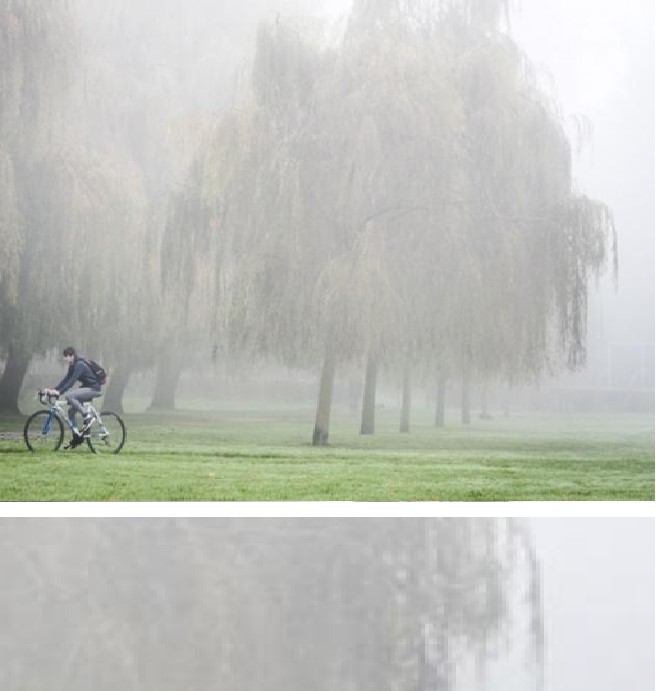} &\hspace{-2.5mm}

\includegraphics[width = 0.1\linewidth]{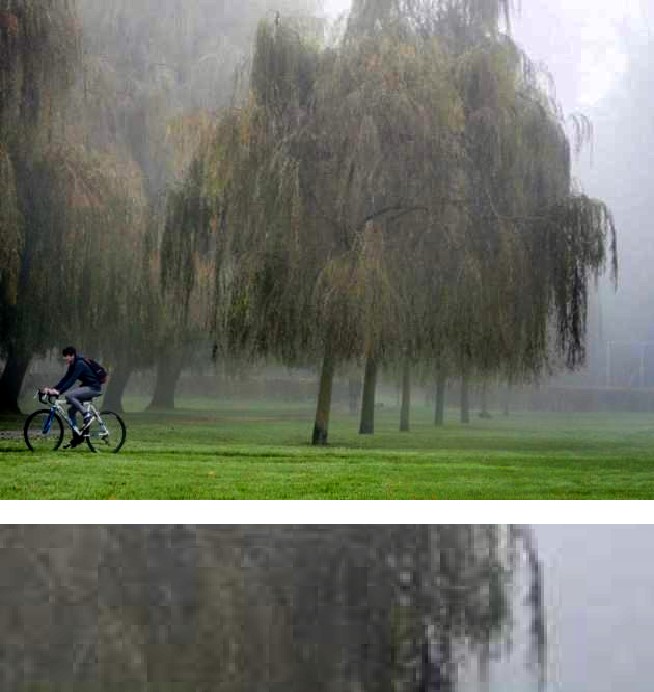} &\hspace{-2.5mm}

\includegraphics[width = 0.1\linewidth]{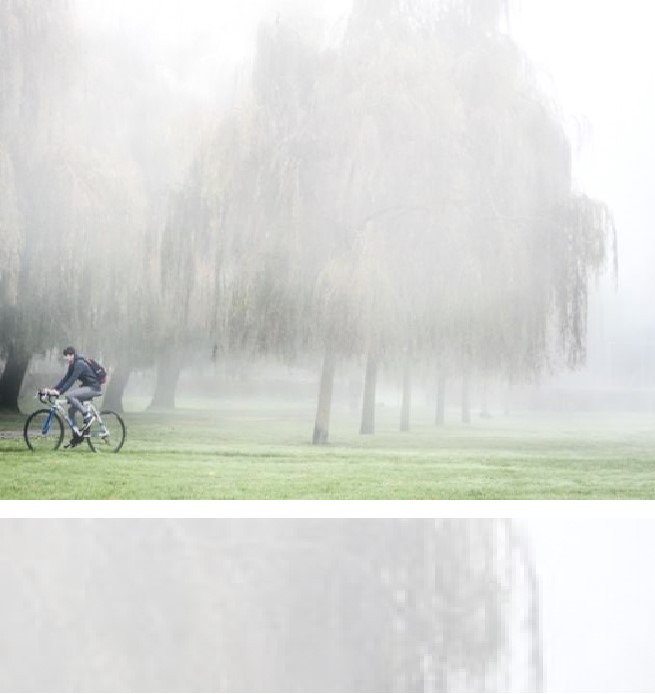} &\hspace{-2.5mm}

\includegraphics[width = 0.1\linewidth]{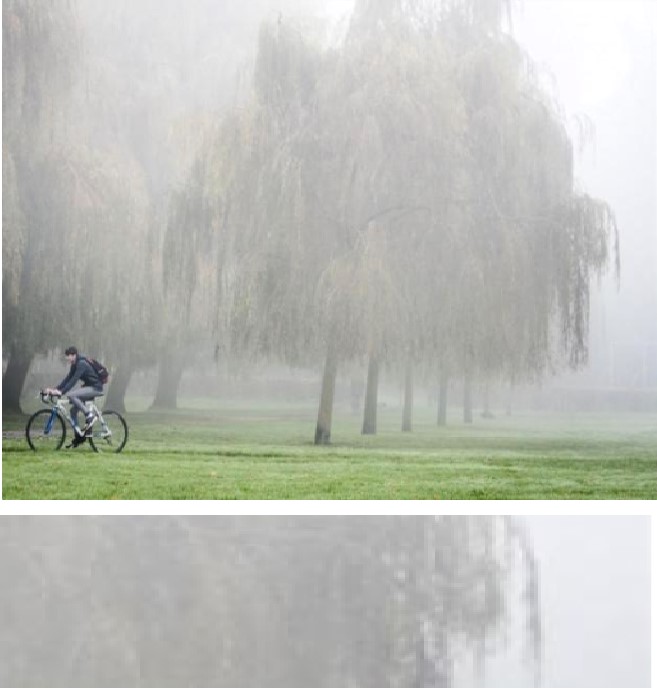} &\hspace{-2.5mm}
\includegraphics[width = 0.1\linewidth]{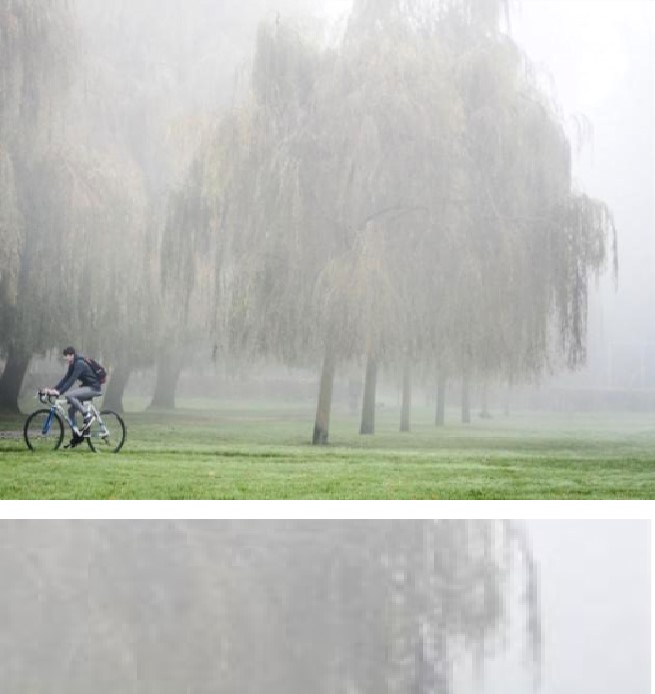} &\hspace{-2.5mm}
\includegraphics[width = 0.1\linewidth]{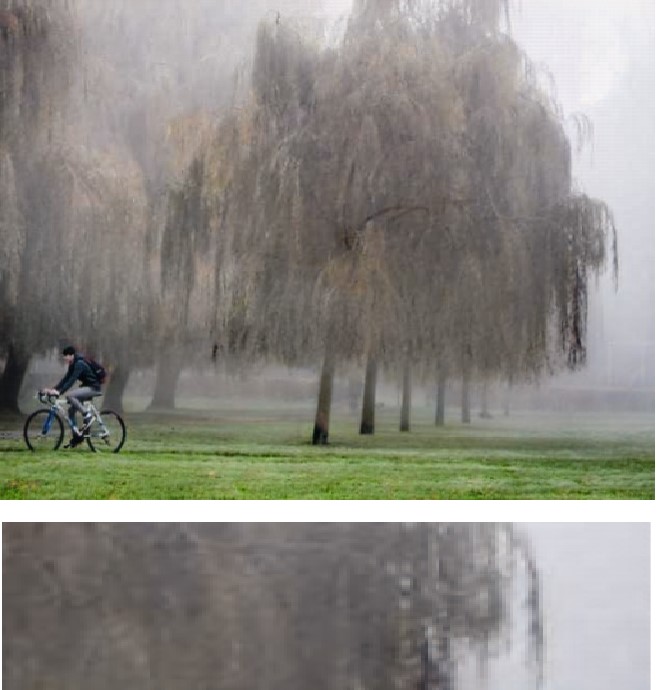} &\hspace{-2.5mm}
\includegraphics[width = 0.1\linewidth]{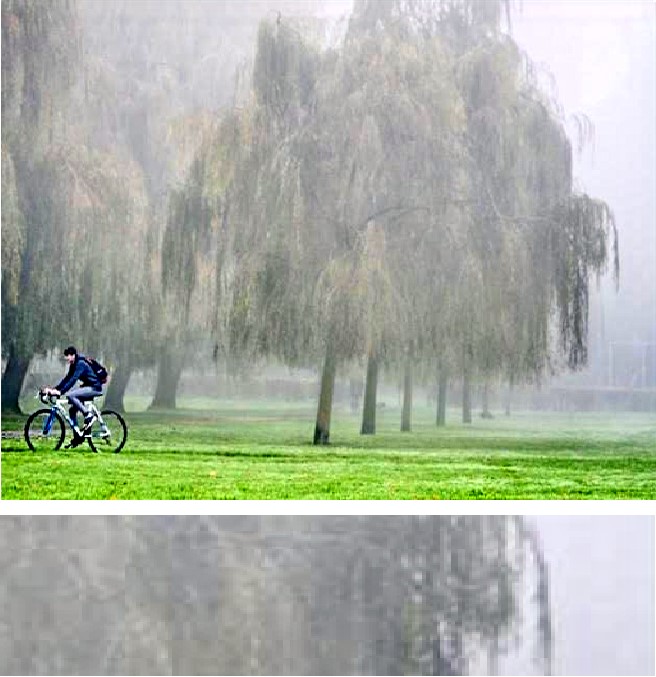} &\hspace{-2.5mm}

\includegraphics[width = 0.1\linewidth]{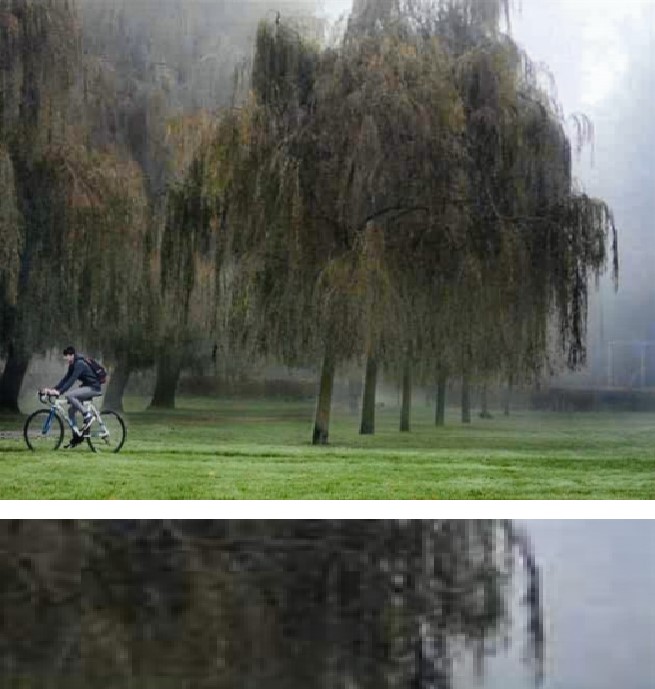} &\hspace{-2.5mm}

\includegraphics[width = 0.1\linewidth]{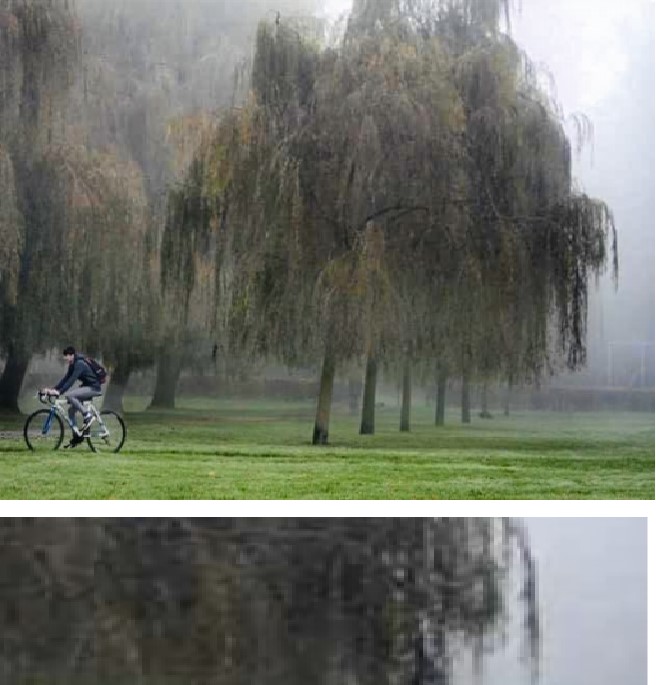} &\hspace{-2.5mm}

\vspace{1mm}\\
% \includegraphics[width = 0.1\linewidth]{samples/Figures/RealComparison4/Original.jpg} &\hspace{-2.5mm}

% \includegraphics[width = 0.1\linewidth]{samples/Figures/RealComparison4/dehazenet.jpg} &\hspace{-2.5mm}

% \includegraphics[width = 0.1\linewidth]{samples/Figures/RealComparison4/FFA.jpg}&\hspace{-2.5mm}
% \includegraphics[width = 0.1\linewidth]{samples/Figures/RealComparison4/MSBDN.jpg} &\hspace{-2.5mm}
% \includegraphics[width = 0.1\linewidth]{samples/Figures/RealComparison4/DA.jpg} &\hspace{-2.5mm}
% \includegraphics[width = 0.1\linewidth]{samples/Figures/RealComparison4/Ours.jpg} &\hspace{-2.5mm}
% &\vspace{-4mm}
% \\
\includegraphics[width = 0.1\linewidth]{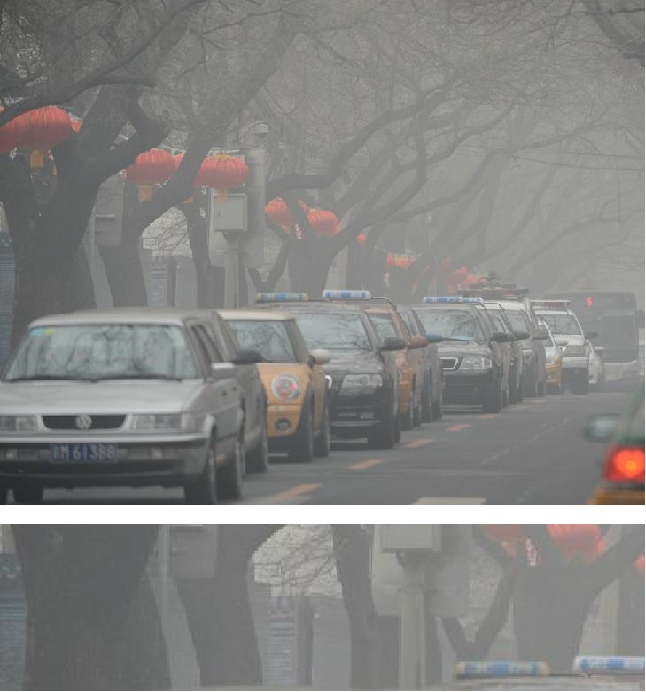} &\hspace{-2.5mm}

\includegraphics[width = 0.1\linewidth]{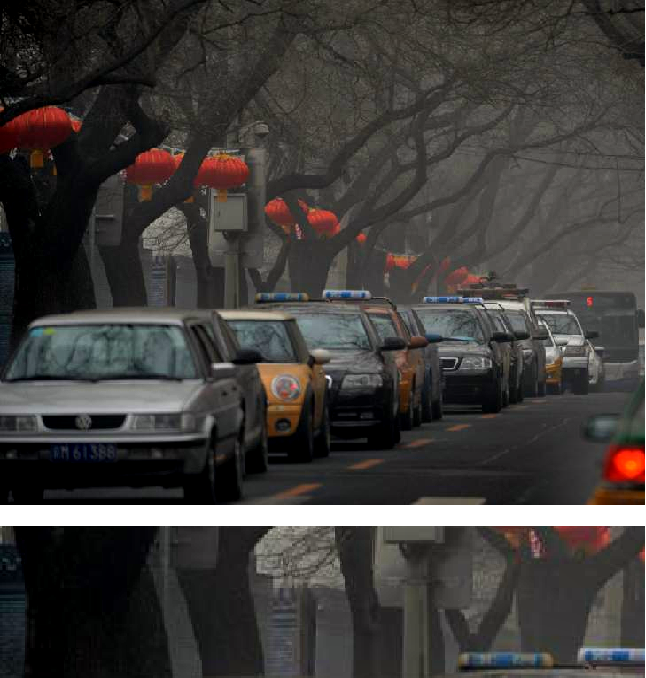} &\hspace{-2.5mm}

\includegraphics[width = 0.1\linewidth]{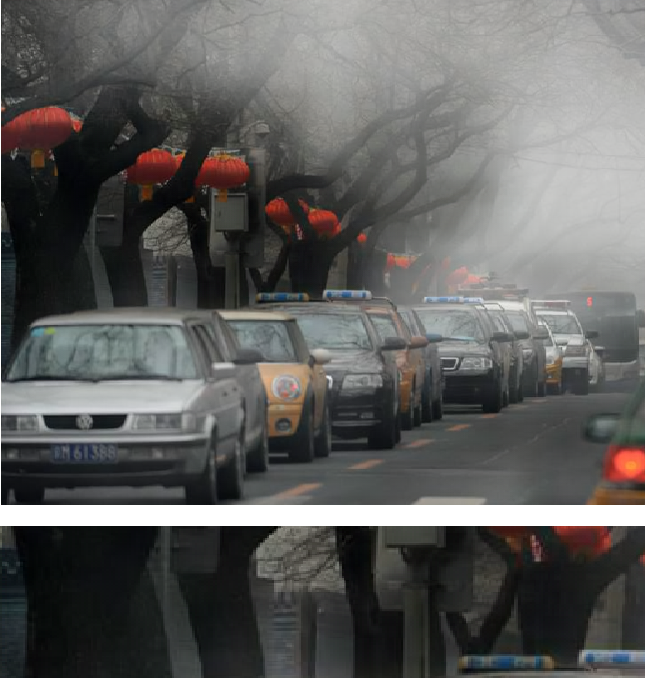} &\hspace{-2.5mm}

\includegraphics[width = 0.1\linewidth]{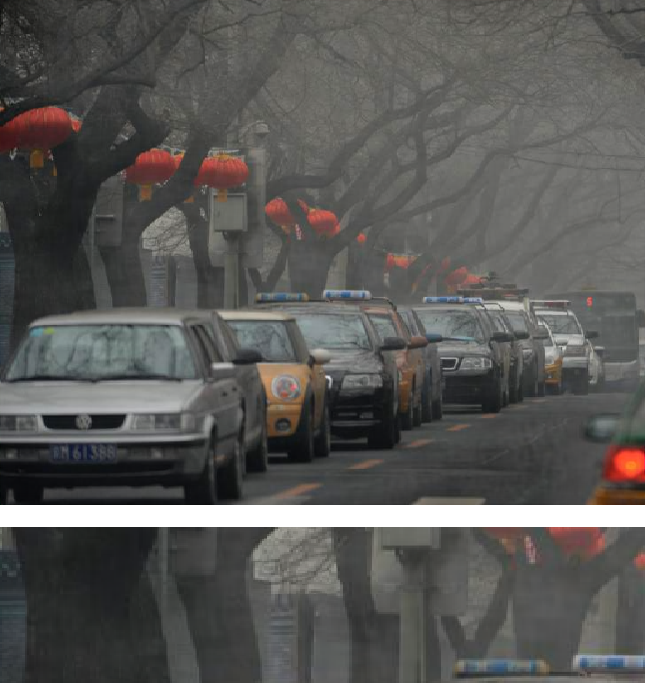}&\hspace{-2.5mm}
\includegraphics[width = 0.1\linewidth]{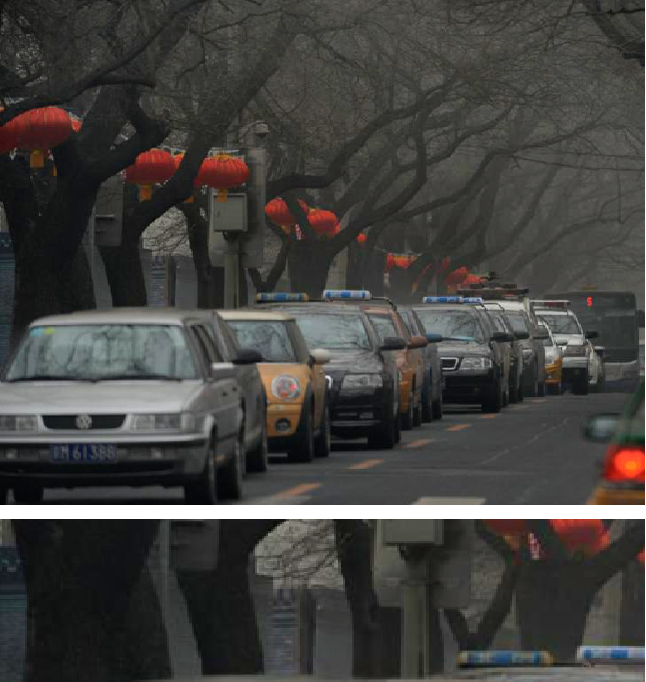} &\hspace{-2.5mm}
\includegraphics[width = 0.1\linewidth]{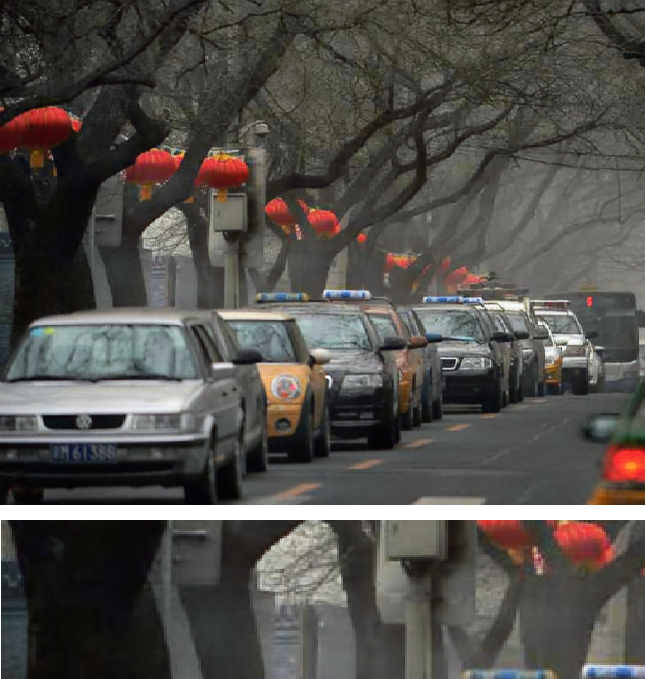} &\hspace{-2.5mm}

\includegraphics[width = 0.1\linewidth]{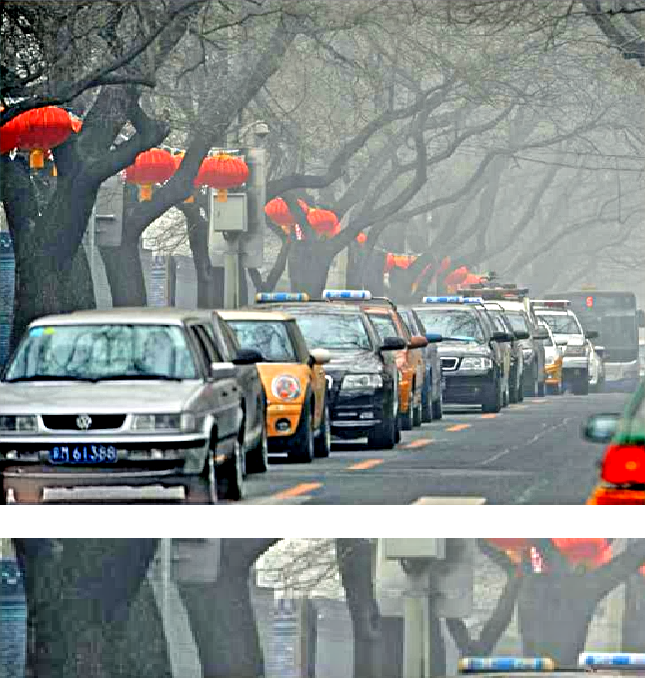} &\hspace{-2.5mm}

\includegraphics[width = 0.1\linewidth]{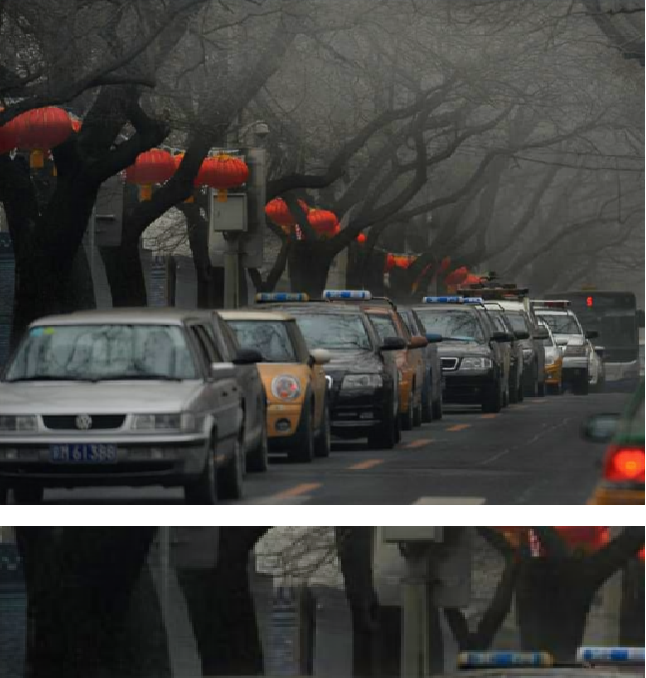} &\hspace{-2.5mm}
\includegraphics[width = 0.1\linewidth]{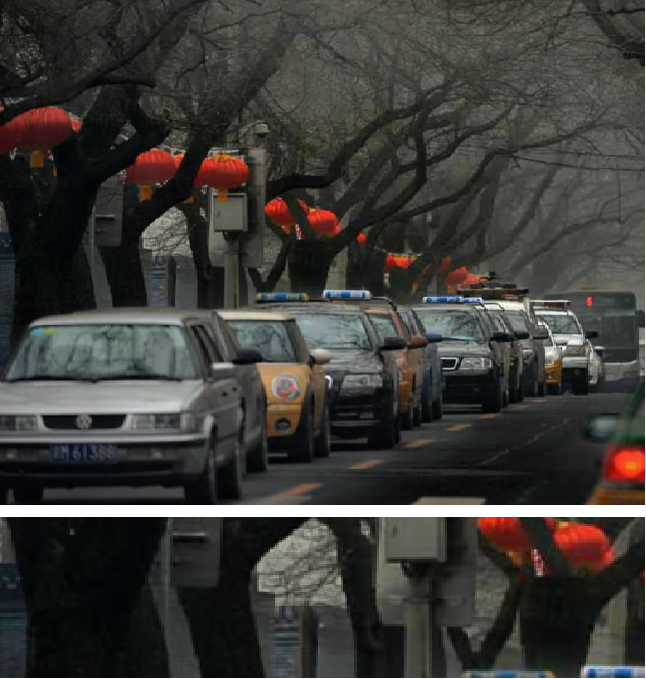} &\hspace{-2.5mm}
\vspace{1mm}\\
\includegraphics[width = 0.1\linewidth]{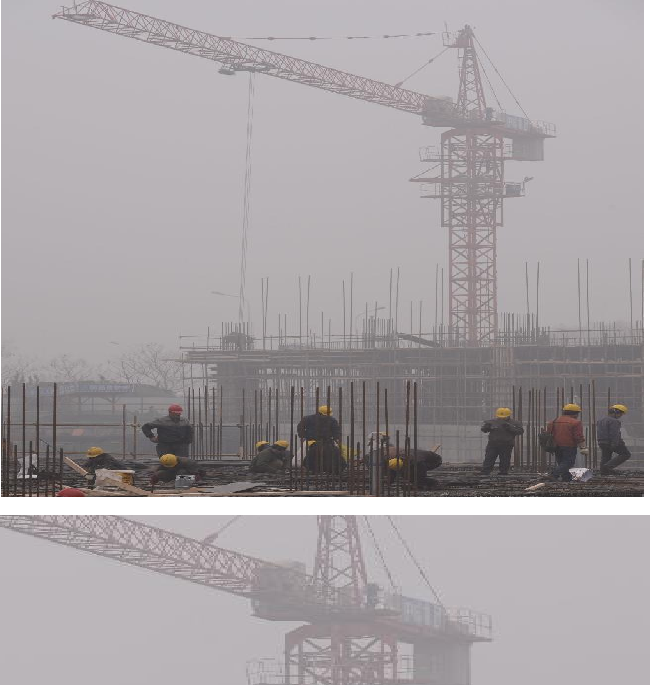} &\hspace{-2.5mm}

\includegraphics[width = 0.1\linewidth]{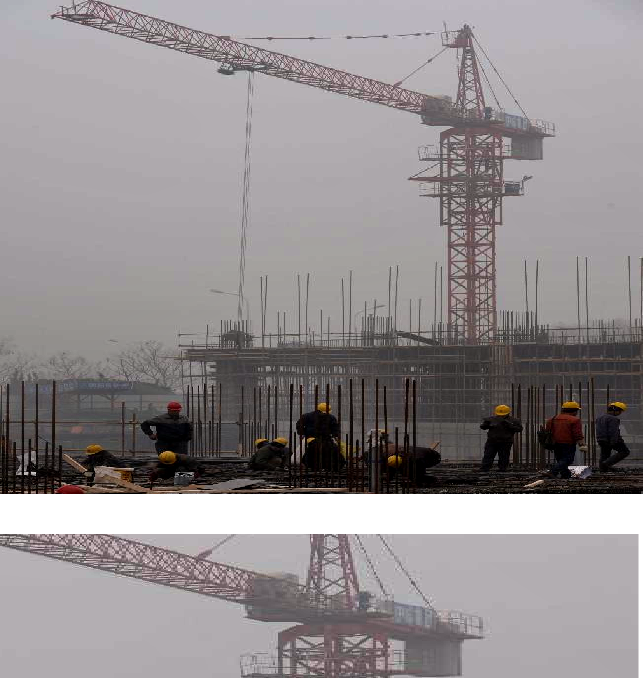}&\hspace{-2.5mm}

\includegraphics[width = 0.1\linewidth]{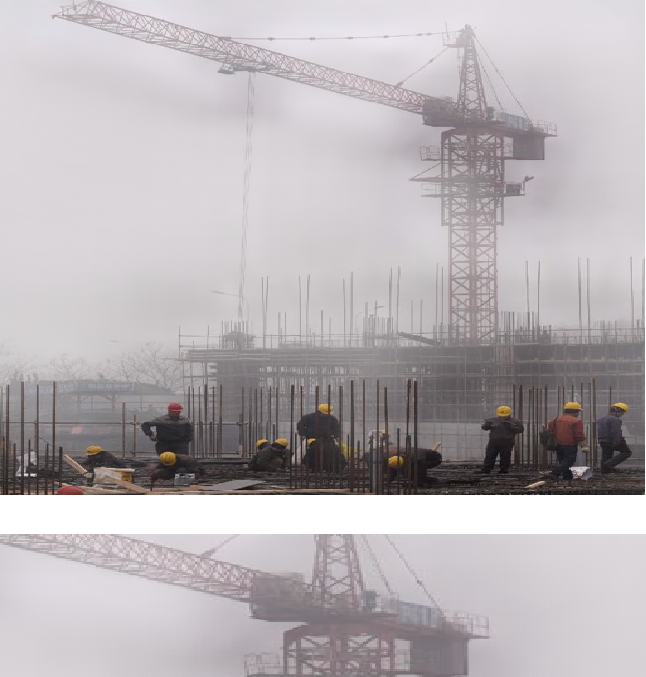} &\hspace{-2.5mm}

\includegraphics[width = 0.1\linewidth]{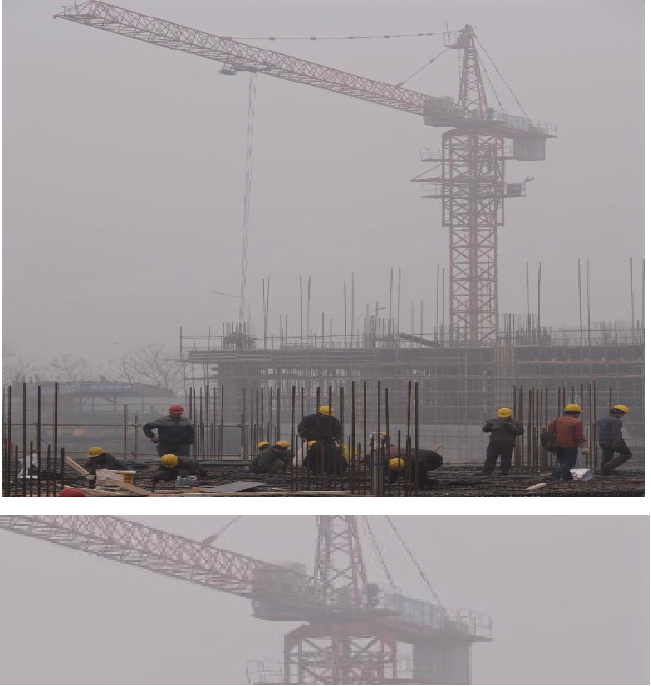}&\hspace{-2.5mm}
\includegraphics[width = 0.1\linewidth]{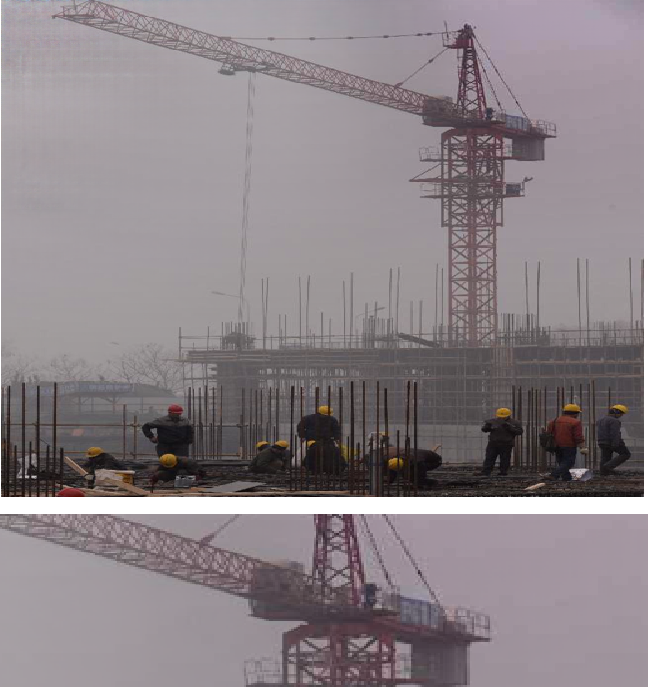} &\hspace{-2.5mm}
\includegraphics[width = 0.1\linewidth]{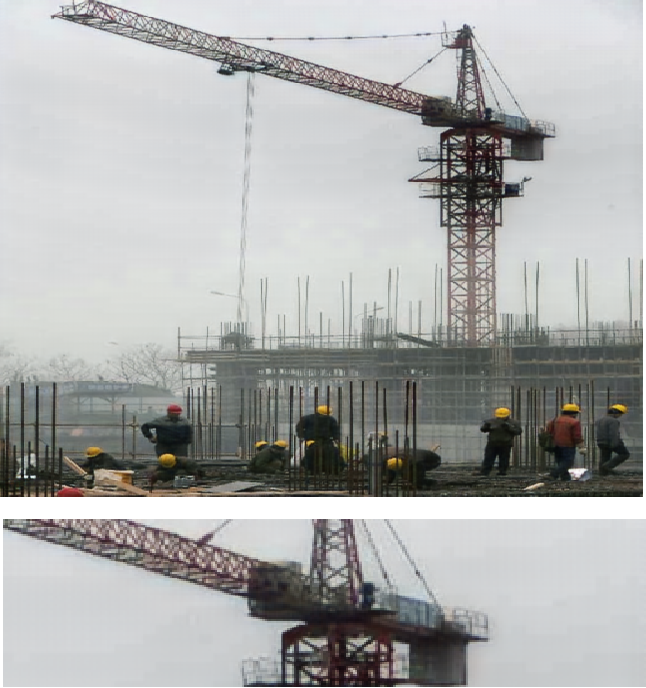} &\hspace{-2.5mm}

\includegraphics[width = 0.1\linewidth]{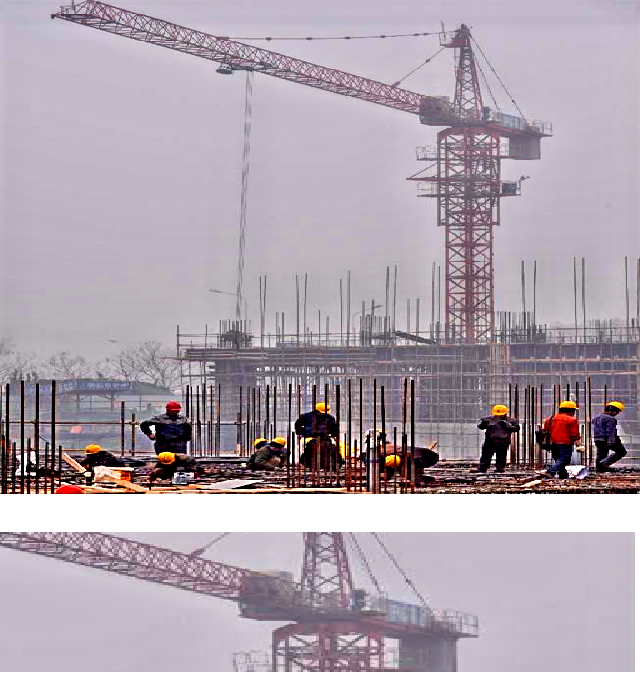} &\hspace{-2.5mm}

\includegraphics[width = 0.1\linewidth]{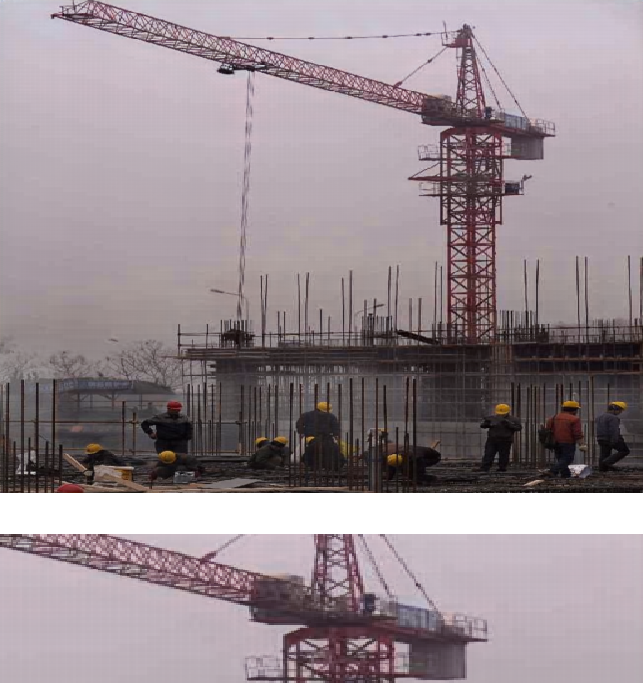} &\hspace{-2.5mm}

\includegraphics[width = 0.1\linewidth]{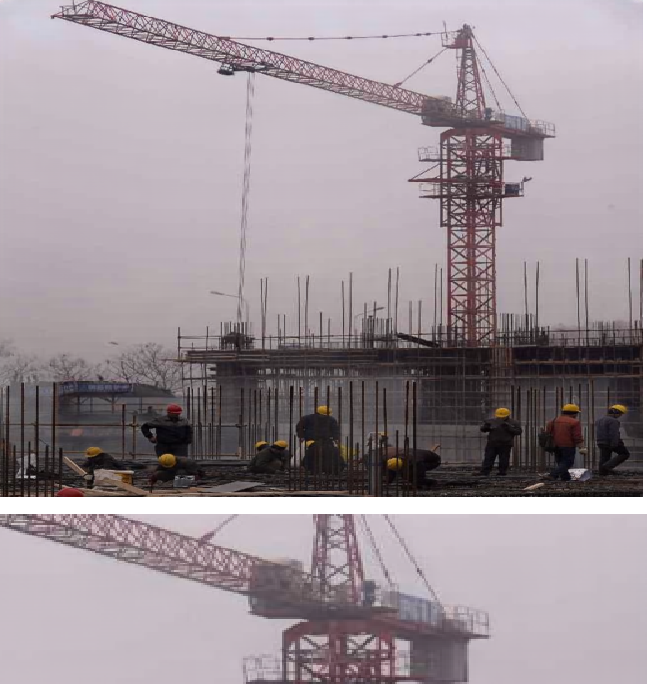} &\hspace{-2.5mm}

\\

\small{(a) Input}  &\hspace{-4mm}  
\small{(b)DehazeNet~\cite{cai2016dehazenet}} &\hspace{-4mm} 
\small{(c) GDN~\cite{griddehazenet}} &\hspace{-4mm}  
\small{(d)FFA-Net~\cite{ffa-net}} &\hspace{-4mm} 
\small{(e)MSBDN~\cite{msbdn}}   &\hspace{-4mm} 
\small{(f)DA~\cite{shao2020domain}} &\hspace{-4mm}
\small{(g)PSD-FFA~\cite{chen2021psd}}  &\hspace{-4mm} 
\small{(h)DAN-Tiny(Ours)}  &\hspace{-4mm} 
\small{(i)DAN(Ours)}  &\hspace{-4mm} 
\\
        \end{tabular}
    \end{center}
\caption{\small{Visual comparisons on real-world hazy images with SOTA dehazing methods(b-g). Our DAN-Net(i) and Tiny DAN-Net(h) also perform well in non-winter haze scenes with uneven degradation.}}\label{fig:visualcomparison3}
\end{figure*}

~\paragraph{Visual Comparison.}
We compare our DAN-Net and DAN-Net-Tiny with SOTA image dehazing or image desnowing methods on the quality of restored images, which are presented in Fig. ~\ref{fig:visualcomparison1}, Fig. ~\ref{fig:visualcomparison2} and Fig. ~\ref{fig:visualcomparison3}. Obviously, our method generates the most natural images, compared to other methods. DAN-Net restores the area covered by the snow streaks perfectly in both synthetic and real winter images, instead of retaining the vestige of snow spots in HDCW-Net and JSTASR in the detailed enlarged images. It's worth mentioning that despite only using synthetic data for training, our method shows superior generalization ability on real-world images. Please refer to our supplementary material for more visual comparisons of synthetic and real-world hazy images.

~\paragraph{Comparison of Computational Complexity.}
For real-time deployment of models, computational complexity is an important aspect to consider for us. To check the computational complexity, we summarize the number of GMacs and parameters of previous state-of-the-art desnowing and dehazing networks in Table~\ref{tab:ComparisonDesnowing},\ref{tab:ComparisonDehazing}. We note that the tiny desnowing expert network we proposed achieves the best parameter-performance trade-off compared to the latest state-of-the-art desnowing method HDCW-Net (ICCV'21). The number of GMacs is only $3.06G$ when the input size of $256\times256$. Notably, FFA-Net and DMT-Net are highly computationally complex even though they are recent state-of-the-art dehazing methods.

~\paragraph{Quantifiable Performance for High-level Vision Tasks.} As shown in Fig.~\ref{fig:GoogleCVAPI}, we show subjective but quantifiable results for a concrete demonstration, in which the labels and corresponding confidences are both supported by Google Vision API\footnote{Google Vision API : \url{https://cloud.google.com/vision/}}. Our comparison illustrates that these bad weather scenes could impede the performance and confidence of high-level vision tasks.

\vspace{0em}
\subsection{Real-world Winter Scenes Dataset (RWSD)}

Considering the lack of real winter scenes dataset for practical testing, we created a Real-world Winter Scenes Dataset (\textbf{RWSD}), which contains 500 real-world winter scene images snapped in complex snowy or hazy scenes. Furthermore, we apply our manner to a part of images from RWSD, as shown in Fig.\ref{fig:realisticrestoration},\ref{fig:GoogleCVAPI}(d). We will release RWSD with code on our project website on GitHub together after the submitted paper is received. Please refer to our supplemental material for more examples of restoration results and corresponding original degraded images from our RWSD.

\subsection{Ablation Study}

\begin{table}
    %\vspace{-0.1cm}
    % \setlength{\abovecaptionskip}{0cm} %调整caption与图的距离
    % \setlength{\belowcaptionskip}{-0.4cm}%调整caption与下文的距离

\caption{\small{Comparing the results of different Multi-branch Spectral Transform Block configurations.}}
\resizebox{8cm}{!}{
\begin{tabular}{cccccc}
\toprule
\textbf{Metric}    & \textbf{VC} & \textbf{LB} & \textbf{GB} & \textbf{MSTBwoST} & \textbf{Ours} \\ \hline\hline
\textbf{Desnowing PSNR/SSIM} & \textbf{27.13/0.91}   & \textbf{27.43/0.92}   & \textbf{27.92/0.91}   & \textbf{28.21/0.92}        & \textbf{30.56 /0.95}      \\ 

\textbf{Dehazing PSNR/SSIM} & \textbf{26.53/0.93}   & \textbf{26.69/0.93}   & \textbf{26.73/0.93}   & \textbf{27.05/0.94}        & \textbf{29.12/0.97}      \\ 
\bottomrule
\end{tabular}}
\label{tab:abalation1}
\end{table}

\begin{table}
%\vspace{.0cm}
\setlength{\abovecaptionskip}{0.0cm} %调整caption与图的距离
\setlength{\belowcaptionskip}{-0.4cm}%调整caption与下文的距离

\caption{\small{Verfication for effectiveness of proposed loss functions.}}
\resizebox{8cm}{!}{
\begin{tabular}{cccccc}
\toprule
\textbf{Metric}    & \textbf{Baseline} & \textbf{$\mathcal{L}_{char}$} &  \textbf{$\mathcal{L}_{char}+\mathcal{L}_{per}$}& \textbf{Ours} \\ \hline\hline
\textbf{Desnowing PSNR/SSIM}    & \textbf{29.26 /0.92}   & \textbf{29.44 /0.92}   & \textbf{30.13 /0.93}        & \textbf{30.56 /0.95}      \\ 
\textbf{Dehazing PSNR/SSIM}   & \textbf{28.42/0.94}   & \textbf{28.56/0.95}   & \textbf{28.73/0.96}        & \textbf{29.12/0.97}      \\
\bottomrule
\end{tabular}}
\label{tab:abalation2}
\end{table}

\begin{table}
%\vspace{-1.0cm}
\setlength{\abovecaptionskip}{0cm} %调整caption与图的距离
\setlength{\belowcaptionskip}{-0.4cm}%调整caption与下文的距离

\caption{\small{Ablation study of the proposed Dual-Pool Attention.}}
\resizebox{8cm}{!}{
\begin{tabular}{cccccc}
\toprule
\textbf{Metric}    & \textbf{wo DP-Att} & \textbf{CA} &  \textbf{PA}& \textbf{Ours} \\ \hline\hline
\textbf{Desnowing PSNR/SSIM} & \textbf{28.76 /0.93}   & \textbf{29.31 /0.94}   & \textbf{29.67 /0.94}      & \textbf{30.56 /0.95}      \\ 
\textbf{Dehazing PSNR/SSIM} & \textbf{28.15/0.94}   & \textbf{28.57/0.96}   & \textbf{28.68/0.96}     & \textbf{29.12/0.97}      \\
\bottomrule
\end{tabular}}
\label{tab:abalation3}
\end{table}

% \begin{table}
% \setlength{\abovecaptionskip}{0.0cm} %调整caption与图的距离
% \setlength{\belowcaptionskip}{0cm}%调整caption与下文的距离
% \caption{Ablation study of the proposed Cross-layer Activation Gated Attention Module.}
% \resizebox{8cm}{!}{
% \begin{tabular}{cccccc}
% \toprule
% \textbf{Metric}    & \textbf{wo DP-Att} & \textbf{CA} &  \textbf{PA}& \textbf{Ours} \\ \hline\hline
% \textbf{Desnowing PSNR/SSIM} & \textbf{28.76 /0.93}   & \textbf{29.31 /0.94}   & \textbf{29.67 /0.94}      & \textbf{30.56 /0.95}      \\ 
% \textbf{Dehazing PSNR/SSIM} & \textbf{28.15/0.94}   & \textbf{28.57/0.96}   & \textbf{28.72/0.96}     & \textbf{29.12/0.97}      \\
% \bottomrule
% \end{tabular}}
% \label{tab:abalation4}
% \end{table}

For a reliable ablation study, we utilize the latest desnowing and dehazing dataset,\textit{i,e.}, CSD~\cite{chen2021all} and Haze4k~\cite{liu2021synthetic}, as the benchmark for training and testing, and choose the expert network to perform our experiments of following ablation studies, 

~\paragraph{Effectiveness of Multi-branch Spectral Transform Block.} 
To prove the superior performance of spectral transform for network training, we replace the Global branch with plain Conv-ELU-Conv branch in the proposed network. Specifically, in Table~\ref{tab:abalation1}, we adopt the vanilla convolution with kernel size $3\times3$ (\textbf{VC}), the Local-branch (\textbf{LB}), the Global-branch (\textbf{GB}), the Multi-scale Spectral Transform Block without Spectral Transform operation (\textbf{MSTBwoST}) and the Multi-branch Spectral Transform Block (\textbf{Ours}). From Table~\ref{tab:abalation1}, we can find that using the Spectral Transform for the degradation feature extraction technique can achieve much better recovered results compared with other configurations.

~\paragraph{Effectiveness of loss functions.}
We evaluate the effectiveness of the proposed loss function $\mathcal{L}_{st}$. Moreover, we compared the $\mathcal{L}_{st}$ with the perceptual loss function $\mathcal{L}_{per}$. The detail implement of perceptual loss is following the previous desnowing method~\cite{chen2021all} in our experiments of this ablation study. The comparison is demonstrated in Table~\ref{tab:abalation2}. It's worth noting that '\textbf{Baseline}' indicates the $L1$ loss on the final recovered result of Table~\ref{tab:abalation2}. The results indicate that with the proposed Spectral Transform loss function, the performance of the network can be improved greatly compared with other configurations of loss function,\textit{i,e.}, $\mathcal{L}_{char}$ and $\mathcal{L}_{char}+L_{per}$.

~\paragraph{Effectiveness of Dual-Pool Attention.}
To verify the effectiveness of the proposed Dual-Pool Attention module, we construct several settings for comparison, that is, (i) The proposed expert network without the Dual-Pool Attention (\textbf{wo DP-Att}). (ii) Remove the Channel Attention Part (\textbf{CA}) of the attention module in the proposed Dual-Pool Attention Module. (iii) Remove the Spatial Attention Part (\textbf{CA}) of the attention module in the proposed Dual-Pool Attention Module. The results indicate that with complete attention mechanism we proposed, the performance of the network can be improved greatly compared with the baseline (\textit{i.e.},\textbf{wo DP-Att}).

Please refer to our supplementary material for more ablation studies of the proposed method.

% ~\paragraph{Parameters Analysis}

\section{Conclusions}
In this paper, a novel degradation-adaptive neural network to address the limitations in current image restoration methods in winter scenes was proposed. First, to handle the complicated snowy or hazy scenario, a compact network as the task-specific expert was developed. With the impressive restoration performance of dehazing expert or desnowing expert, the complex degradation of haze or snow can be removed effectively. More importantly, we develop a novel Adaptive Gated Neural inspired by MoE strategy. Based on it, the DAN-Net adaptively handle the complex winter degradation by haze and snow. Experimental results showed that the proposed method outperforms state-of-the-art dehazing or desnowing algorithms and can benefit high-level vision tasks.

\begin{figure*}[h]
    \centering
    \includegraphics[width=0.9\textwidth]{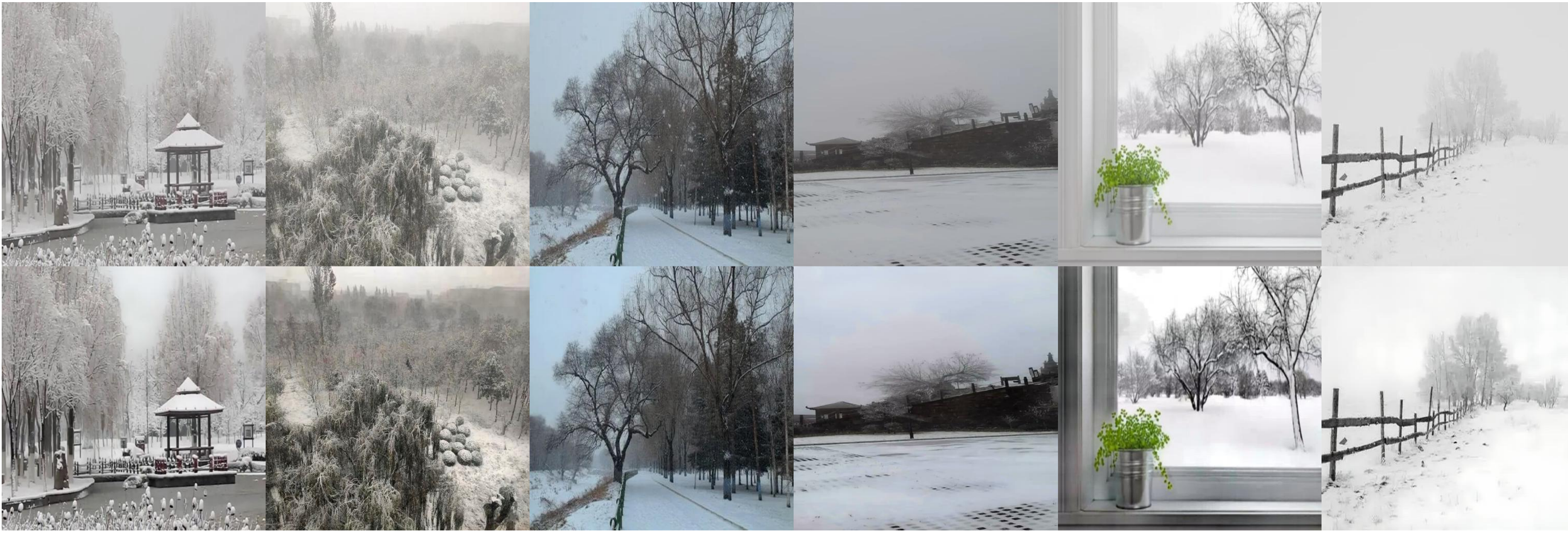}
    \caption{Realistic winter images (top) from the RWSD dataset created by this work and corresponding restoration results (bottom) using the proposed Degradation-Adaptive Neural Network. Zoom-in for better visual quality.}
    \label{fig:realisticrestoration1}
\end{figure*}

\begin{figure*}[h]
    \centering
    \includegraphics[width=0.9\textwidth]{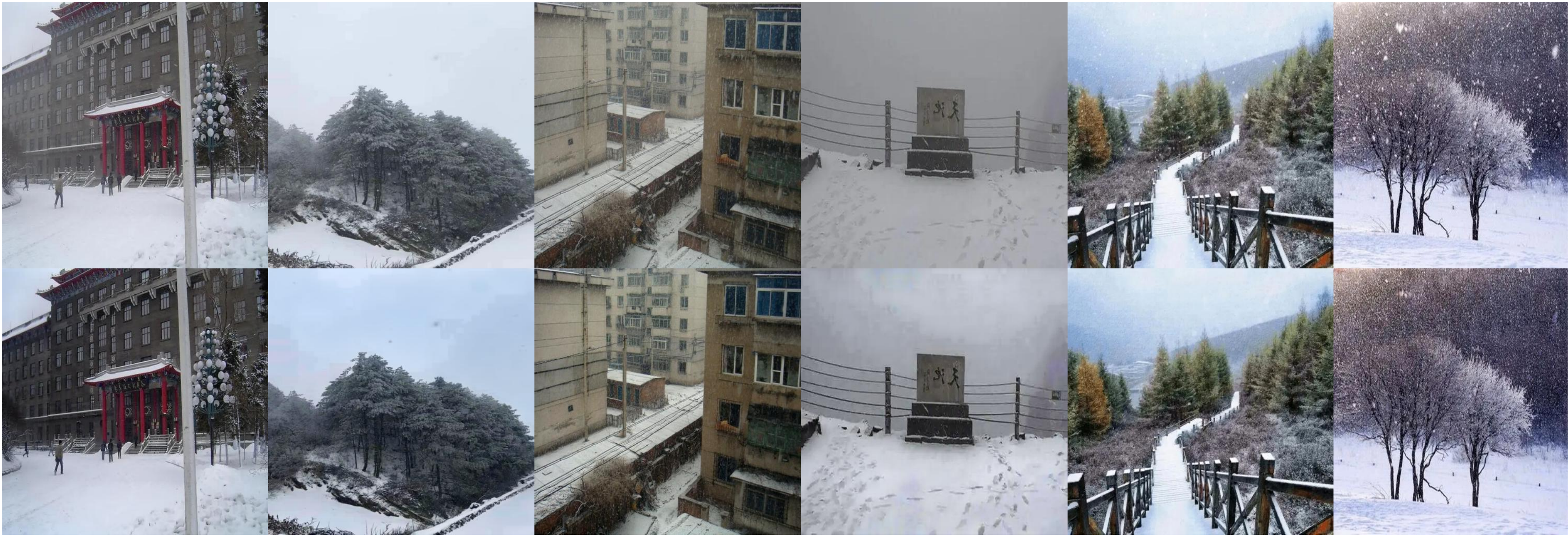}
    \caption{Realistic winter images (top) from the RWSD dataset created by this work and corresponding restoration results (bottom) using the proposed Degradation-Adaptive Neural Network. Zoom-in for better visual quality.}
    \label{fig:realisticrestoration2}
\end{figure*}

\begin{figure*}[h]
    \centering
    \includegraphics[width=0.9\textwidth]{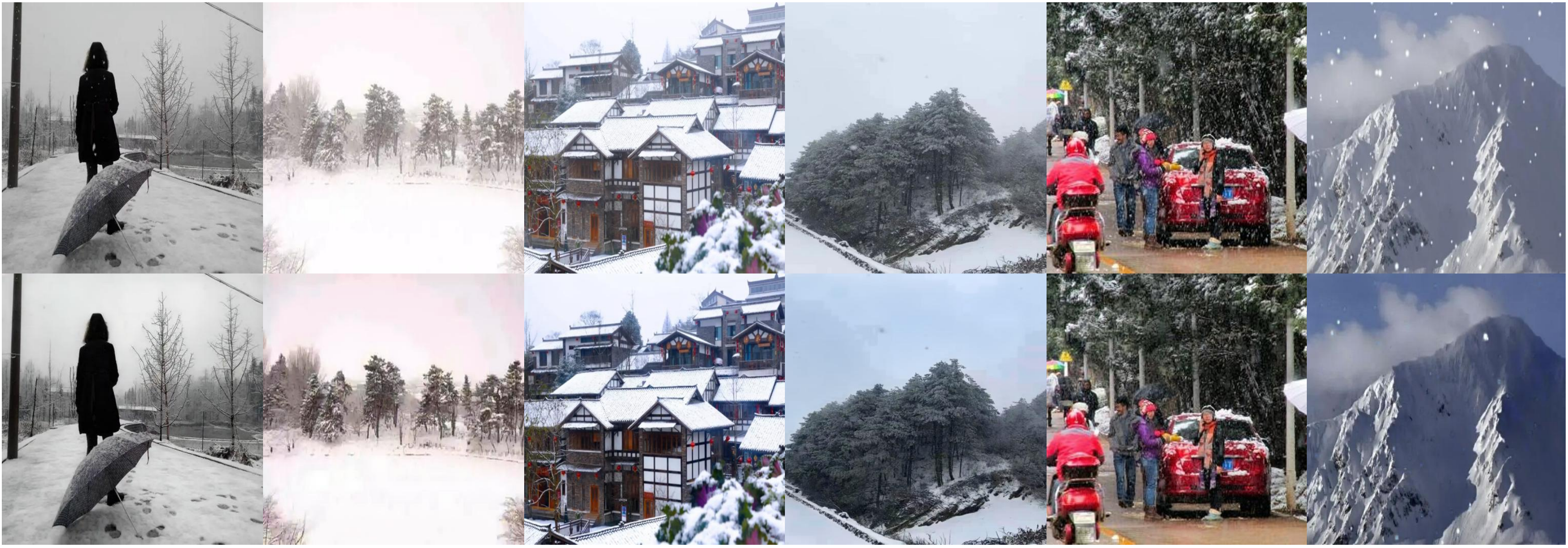}
    \caption{Realistic winter images (top) from the RWSD dataset created by this work and corresponding restoration results (bottom) using the proposed Degradation-Adaptive Neural Network. Zoom-in for better visual quality.}
    \label{fig:realisticrestoration3}
\end{figure*}

\begin{figure*}[h!]
    \vspace{-1mm}
    \setlength{\abovecaptionskip}{0cm} %调整caption与图的距离
\setlength{\belowcaptionskip}{-0.8cm}   %调整图片标题与下文距离
    \begin{center}
        \begin{tabular}{ccccccccc}

\includegraphics[width = 0.11\linewidth]{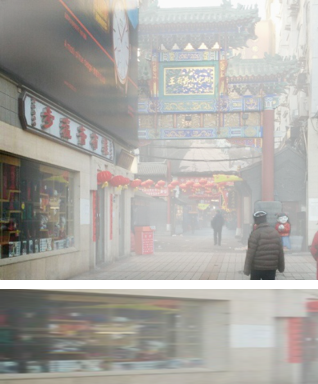} &\hspace{-2.5mm}
\includegraphics[width = 0.11\linewidth]{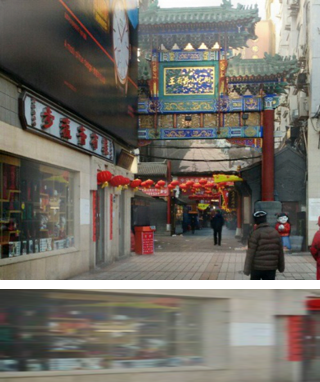} &\hspace{-2.5mm}
\includegraphics[width = 0.11\linewidth]{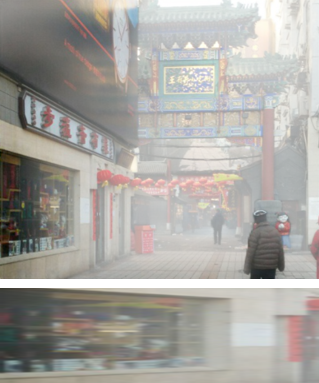} &\hspace{-2.5mm}
\includegraphics[width = 0.11\linewidth]{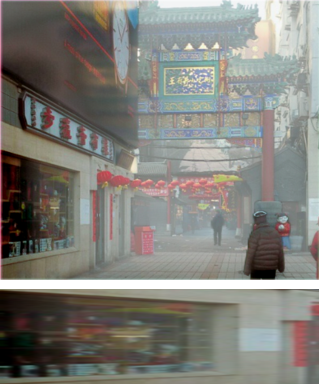} &\hspace{-2.5mm}
\includegraphics[width = 0.11\linewidth]{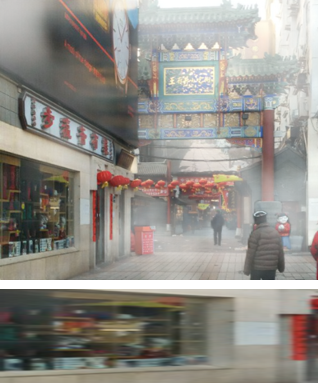} &\hspace{-2.5mm}
\includegraphics[width = 0.11\linewidth]{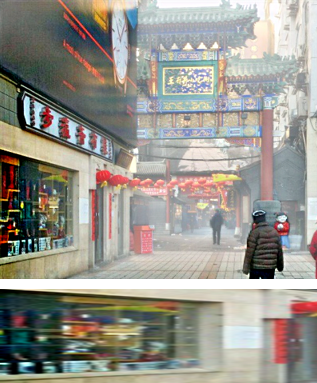} &\hspace{-2.5mm}
\includegraphics[width = 0.11\linewidth]{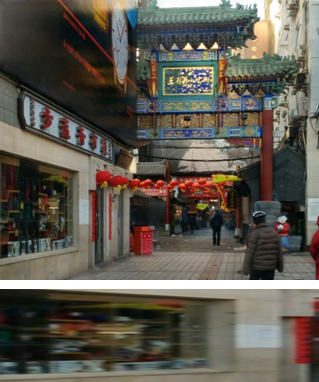} &\hspace{-2.5mm}
\includegraphics[width = 0.11\linewidth]{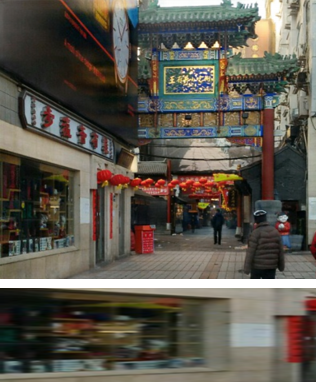} &\hspace{-2.5mm}

\vspace{1mm}\\

\includegraphics[width = 0.11\linewidth]{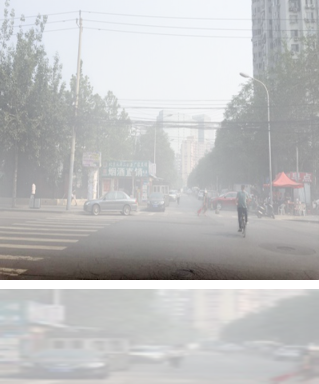} &\hspace{-2.5mm}
\includegraphics[width = 0.11\linewidth]{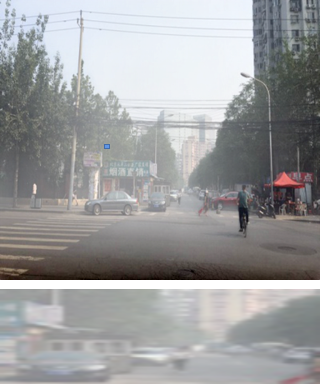} &\hspace{-2.5mm}
\includegraphics[width = 0.11\linewidth]{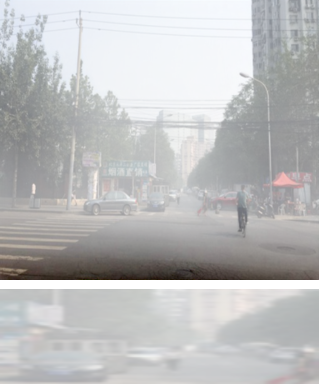} &\hspace{-2.5mm}
\includegraphics[width = 0.11\linewidth]{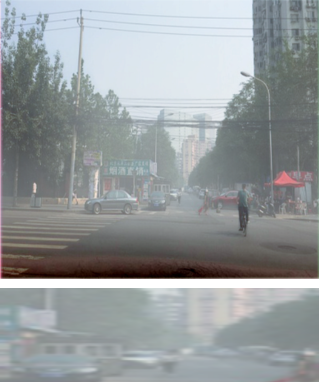} &\hspace{-2.5mm}
\includegraphics[width = 0.11\linewidth]{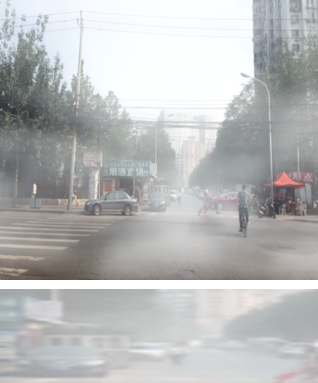} &\hspace{-2.5mm}
\includegraphics[width = 0.11\linewidth]{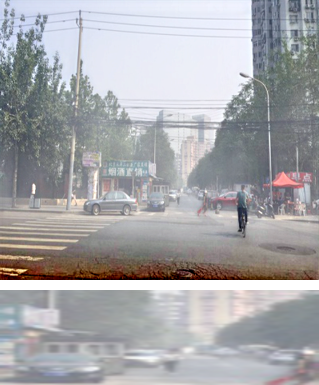} &\hspace{-2.5mm}
\includegraphics[width = 0.11\linewidth]{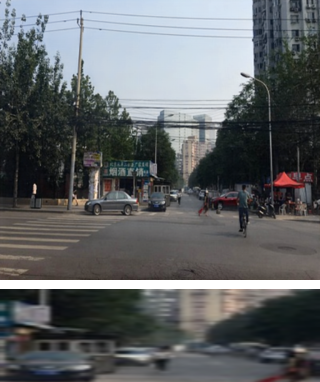} &\hspace{-2.5mm}
\includegraphics[width = 0.11\linewidth]{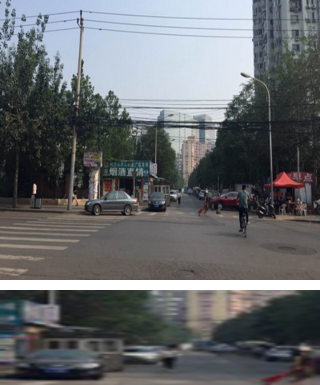} &\hspace{-2.5mm}

\\

\small{(a)Input}  &\hspace{-4mm} 
\small{(b)MSBDN~\cite{msbdn}} &\hspace{-4mm} 
\small{(c) FFA~\cite{ffa-net}} &\hspace{-4mm}  
\small{(d)AOD~\cite{aod}} &\hspace{-4mm}
\small{(e)GDN~\cite{griddehazenet}} &\hspace{-4mm}
\small{(f)PSD~\cite{chen2021psd}} &\hspace{-4mm} 
\small{(g)DAN(Ours)} &\hspace{-4mm} 
\small{(h)Ground-truth} &\hspace{-4mm} 
\\
        \end{tabular}
    \end{center}
\caption{\small{Visual comparisons of results produced by our DAN-Net(g) and SOTA image dehazing methods(b-f) on the hazy synthetic image from the test dataset of Haze4k~\cite{liu2021synthetic} (a).}}\label{fig:visualcomparison_haze1}
\end{figure*}

\begin{figure*}[h!]
     \vspace{6mm}
    \setlength{\abovecaptionskip}{0cm} %调整caption与图的距离
\setlength{\belowcaptionskip}{-0.5cm}   %调整图片标题与下文距离
    \begin{center}
        \begin{tabular}{ccccccccc}

\includegraphics[width = 0.11\linewidth]{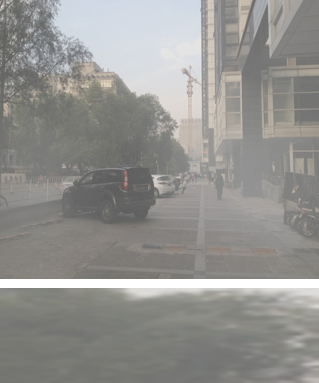} &\hspace{-2.5mm}
\includegraphics[width = 0.11\linewidth]{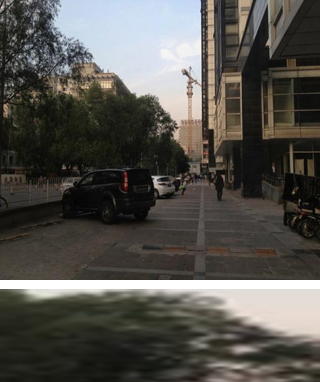} &\hspace{-2.5mm}
\includegraphics[width = 0.11\linewidth]{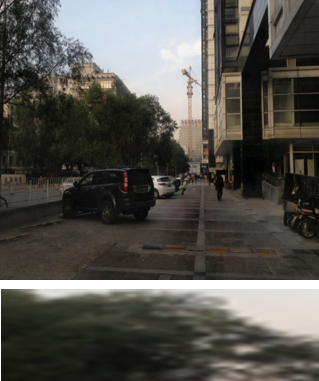} &\hspace{-2.5mm}
\includegraphics[width = 0.11\linewidth]{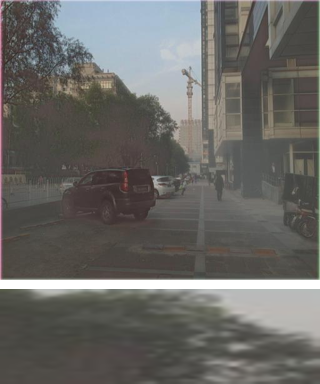} &\hspace{-2.5mm}
\includegraphics[width = 0.11\linewidth]{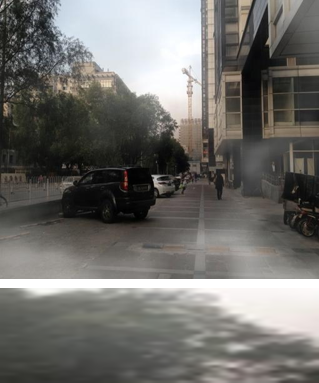} &\hspace{-2.5mm}
\includegraphics[width = 0.11\linewidth]{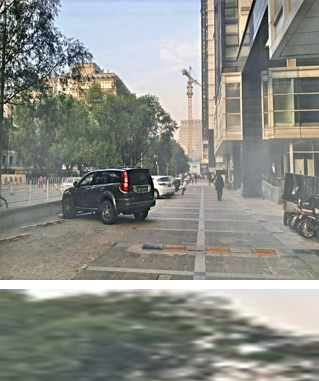} &\hspace{-2.5mm}
\includegraphics[width = 0.11\linewidth]{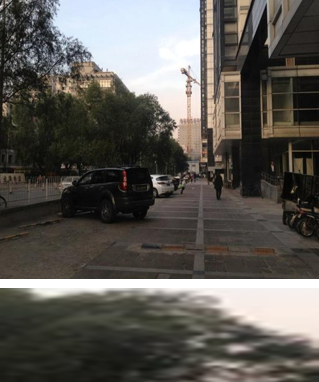} &\hspace{-2.5mm}
\includegraphics[width = 0.11\linewidth]{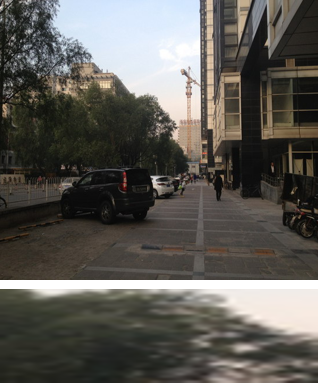} &\hspace{-2.5mm}

\\

\small{(a)Input}  &\hspace{-4mm} 
\small{(b)MSBDN~\cite{msbdn}} &\hspace{-4mm} 
\small{(c) FFA~\cite{ffa-net}} &\hspace{-4mm}  
\small{(d)AOD~\cite{aod}} &\hspace{-4mm}
\small{(e)GDN~\cite{griddehazenet}} &\hspace{-4mm}
\small{(f)PSD~\cite{chen2021psd}} &\hspace{-4mm} 
\small{(g)DAN(Ours)} &\hspace{-4mm} 
\small{(h)Ground-truth} &\hspace{-4mm} 
\\
        \end{tabular}
    \end{center}
\caption{\small{Visual comparisons of results produced by our DAN-Net(g) and SOTA image dehazing methods(b-f) on hazy synthetic images from SOTS~\cite{SOTS} Outdoor test dataset(a).}}\label{fig:visualcomparison_haze2}
\end{figure*}

\bibliographystyle{ACM-Reference-Format}
\bibliography{sample-sigconf-arxiv}


\begin{thebibliography}{47}

%%% the Web address.

\ifx \showCODEN    \undefined \def \showCODEN     #1{\unskip}     \fi
\ifx \showDOI      \undefined \def \showDOI       #1{#1}\fi
\ifx \showISBNx    \undefined \def \showISBNx     #1{\unskip}     \fi
\ifx \showISBNxiii \undefined \def \showISBNxiii  #1{\unskip}     \fi
\ifx \showISSN     \undefined \def \showISSN      #1{\unskip}     \fi
\ifx \showLCCN     \undefined \def \showLCCN      #1{\unskip}     \fi
\ifx \shownote     \undefined \def \shownote      #1{#1}          \fi
\ifx \showarticletitle \undefined \def \showarticletitle #1{#1}   \fi
\ifx \showURL      \undefined \def \showURL       {\relax}        \fi
% invisible to TeX
\providecommand\bibfield[2]{#2}
\providecommand\bibinfo[2]{#2}
\providecommand\natexlab[1]{#1}
\providecommand\showeprint[2][]{arXiv:#2}

\bibitem[Aljundi et~al\mbox{.}(2017)]%
        {aljundi2017expert}
\bibfield{author}{\bibinfo{person}{Rahaf Aljundi}, \bibinfo{person}{Punarjay
  Chakravarty}, {and} \bibinfo{person}{Tinne Tuytelaars}.}
  \bibinfo{year}{2017}\natexlab{}.
\newblock \showarticletitle{Expert gate: Lifelong learning with a network of
  experts}. In \bibinfo{booktitle}{\emph{Proceedings of the IEEE Conference on
  Computer Vision and Pattern Recognition}}. \bibinfo{pages}{3366--3375}.
\newblock


\bibitem[Berman et~al\mbox{.}(2016)]%
        {berman2016non}
\bibfield{author}{\bibinfo{person}{Dana Berman}, \bibinfo{person}{Shai Avidan},
  {et~al\mbox{.}}} \bibinfo{year}{2016}\natexlab{}.
\newblock \showarticletitle{Non-local image dehazing}. In
  \bibinfo{booktitle}{\emph{Proceedings of the IEEE conference on computer
  vision and pattern recognition}}. \bibinfo{pages}{1674--1682}.
\newblock


\bibitem[Berman et~al\mbox{.}(2017)]%
        {berman2017air}
\bibfield{author}{\bibinfo{person}{Dana Berman}, \bibinfo{person}{Tali
  Treibitz}, {and} \bibinfo{person}{Shai Avidan}.}
  \bibinfo{year}{2017}\natexlab{}.
\newblock \showarticletitle{Air-light estimation using haze-lines}. In
  \bibinfo{booktitle}{\emph{2017 IEEE International Conference on Computational
  Photography (ICCP)}}. IEEE, \bibinfo{pages}{1--9}.
\newblock


\bibitem[Cai et~al\mbox{.}(2016)]%
        {cai2016dehazenet}
\bibfield{author}{\bibinfo{person}{Bolun Cai}, \bibinfo{person}{Xiangmin Xu},
  \bibinfo{person}{Kui Jia}, \bibinfo{person}{Chunmei Qing}, {and}
  \bibinfo{person}{Dacheng Tao}.} \bibinfo{year}{2016}\natexlab{}.
\newblock \showarticletitle{Dehazenet: An end-to-end system for single image
  haze removal}.
\newblock \bibinfo{journal}{\emph{IEEE Transactions on Image Processing}}
  \bibinfo{volume}{25}, \bibinfo{number}{11} (\bibinfo{year}{2016}),
  \bibinfo{pages}{5187--5198}.
\newblock


\bibitem[Charbonnier et~al\mbox{.}(1994)]%
        {charbonnier1994two}
\bibfield{author}{\bibinfo{person}{Pierre Charbonnier}, \bibinfo{person}{Laure
  Blanc-Feraud}, \bibinfo{person}{Gilles Aubert}, {and} \bibinfo{person}{Michel
  Barlaud}.} \bibinfo{year}{1994}\natexlab{}.
\newblock \showarticletitle{Two deterministic half-quadratic regularization
  algorithms for computed imaging}. In \bibinfo{booktitle}{\emph{Proceedings of
  1st International Conference on Image Processing}}, Vol.~\bibinfo{volume}{2}.
  IEEE, \bibinfo{pages}{168--172}.
\newblock


\bibitem[Chen et~al\mbox{.}(2020)]%
        {chen2020jstasr}
\bibfield{author}{\bibinfo{person}{Wei-Ting Chen}, \bibinfo{person}{Hao-Yu
  Fang}, \bibinfo{person}{Jian-Jiun Ding}, \bibinfo{person}{Cheng-Che Tsai},
  {and} \bibinfo{person}{Sy-Yen Kuo}.} \bibinfo{year}{2020}\natexlab{}.
\newblock \showarticletitle{JSTASR: Joint size and transparency-aware snow
  removal algorithm based on modified partial convolution and veiling effect
  removal}. In \bibinfo{booktitle}{\emph{European Conference on Computer
  Vision}}. Springer, \bibinfo{pages}{754--770}.
\newblock


\bibitem[Chen et~al\mbox{.}(2021a)]%
        {chen2021all}
\bibfield{author}{\bibinfo{person}{Wei-Ting Chen}, \bibinfo{person}{Hao-Yu
  Fang}, \bibinfo{person}{Cheng-Lin Hsieh}, \bibinfo{person}{Cheng-Che Tsai},
  \bibinfo{person}{I Chen}, \bibinfo{person}{Jian-Jiun Ding},
  \bibinfo{person}{Sy-Yen Kuo}, {et~al\mbox{.}}}
  \bibinfo{year}{2021}\natexlab{a}.
\newblock \showarticletitle{ALL Snow Removed: Single Image Desnowing Algorithm
  Using Hierarchical Dual-Tree Complex Wavelet Representation and Contradict
  Channel Loss}. In \bibinfo{booktitle}{\emph{Proceedings of the IEEE/CVF
  International Conference on Computer Vision}}. \bibinfo{pages}{4196--4205}.
\newblock


\bibitem[Chen et~al\mbox{.}(2021b)]%
        {chen2021psd}
\bibfield{author}{\bibinfo{person}{Zeyuan Chen}, \bibinfo{person}{Yangchao
  Wang}, \bibinfo{person}{Yang Yang}, {and} \bibinfo{person}{Dong Liu}.}
  \bibinfo{year}{2021}\natexlab{b}.
\newblock \showarticletitle{PSD: Principled synthetic-to-real dehazing guided
  by physical priors}. In \bibinfo{booktitle}{\emph{Proceedings of the IEEE/CVF
  Conference on Computer Vision and Pattern Recognition}}.
  \bibinfo{pages}{7180--7189}.
\newblock


\bibitem[Dong et~al\mbox{.}(2020)]%
        {msbdn}
\bibfield{author}{\bibinfo{person}{Hang Dong}, \bibinfo{person}{Jinshan Pan},
  \bibinfo{person}{Lei Xiang}, \bibinfo{person}{Zhe Hu}, \bibinfo{person}{Xinyi
  Zhang}, \bibinfo{person}{Fei Wang}, {and} \bibinfo{person}{Ming-Hsuan Yang}.}
  \bibinfo{year}{2020}\natexlab{}.
\newblock \showarticletitle{Multi-scale boosted dehazing network with dense
  feature fusion}. In \bibinfo{booktitle}{\emph{Proceedings of the IEEE/CVF
  Conference on Computer Vision and Pattern Recognition}}.
  \bibinfo{pages}{2157--2167}.
\newblock


\bibitem[Engin et~al\mbox{.}(2018)]%
        {engin2018cycle}
\bibfield{author}{\bibinfo{person}{Deniz Engin}, \bibinfo{person}{Anil
  Gen{\c{c}}}, {and} \bibinfo{person}{Hazim Kemal~Ekenel}.}
  \bibinfo{year}{2018}\natexlab{}.
\newblock \showarticletitle{Cycle-dehaze: Enhanced cyclegan for single image
  dehazing}. In \bibinfo{booktitle}{\emph{Proceedings of the IEEE Conference on
  Computer Vision and Pattern Recognition Workshops}}.
  \bibinfo{pages}{825--833}.
\newblock


\bibitem[Fattal(2014)]%
        {fattal2014dehazing}
\bibfield{author}{\bibinfo{person}{Raanan Fattal}.}
  \bibinfo{year}{2014}\natexlab{}.
\newblock \showarticletitle{Dehazing using color-lines}.
\newblock \bibinfo{journal}{\emph{ACM transactions on graphics (TOG)}}
  \bibinfo{volume}{34}, \bibinfo{number}{1} (\bibinfo{year}{2014}),
  \bibinfo{pages}{1--14}.
\newblock


\bibitem[Gross et~al\mbox{.}(2017)]%
        {gross2017hard}
\bibfield{author}{\bibinfo{person}{Sam Gross}, \bibinfo{person}{Marc'Aurelio
  Ranzato}, {and} \bibinfo{person}{Arthur Szlam}.}
  \bibinfo{year}{2017}\natexlab{}.
\newblock \showarticletitle{Hard mixtures of experts for large scale weakly
  supervised vision}. In \bibinfo{booktitle}{\emph{Proceedings of the IEEE
  Conference on Computer Vision and Pattern Recognition}}.
  \bibinfo{pages}{6865--6873}.
\newblock


\bibitem[He et~al\mbox{.}(2021)]%
        {he2021fastmoe}
\bibfield{author}{\bibinfo{person}{Jiaao He}, \bibinfo{person}{Jiezhong Qiu},
  \bibinfo{person}{Aohan Zeng}, \bibinfo{person}{Zhilin Yang},
  \bibinfo{person}{Jidong Zhai}, {and} \bibinfo{person}{Jie Tang}.}
  \bibinfo{year}{2021}\natexlab{}.
\newblock \showarticletitle{Fastmoe: A fast mixture-of-expert training system}.
\newblock \bibinfo{journal}{\emph{arXiv preprint arXiv:2103.13262}}
  (\bibinfo{year}{2021}).
\newblock


\bibitem[He et~al\mbox{.}(2010)]%
        {he2010single}
\bibfield{author}{\bibinfo{person}{Kaiming He}, \bibinfo{person}{Jian Sun},
  {and} \bibinfo{person}{Xiaoou Tang}.} \bibinfo{year}{2010}\natexlab{}.
\newblock \showarticletitle{Single image haze removal using dark channel
  prior}.
\newblock \bibinfo{journal}{\emph{IEEE transactions on pattern analysis and
  machine intelligence}} \bibinfo{volume}{33}, \bibinfo{number}{12}
  (\bibinfo{year}{2010}), \bibinfo{pages}{2341--2353}.
\newblock


\bibitem[Hirzer et~al\mbox{.}(2011)]%
        {hirzer2011person}
\bibfield{author}{\bibinfo{person}{Martin Hirzer}, \bibinfo{person}{Csaba
  Beleznai}, \bibinfo{person}{Peter~M Roth}, {and} \bibinfo{person}{Horst
  Bischof}.} \bibinfo{year}{2011}\natexlab{}.
\newblock \showarticletitle{Person re-identification by descriptive and
  discriminative classification}. In \bibinfo{booktitle}{\emph{Scandinavian
  conference on Image analysis}}. Springer, \bibinfo{pages}{91--102}.
\newblock


\bibitem[Jacobs et~al\mbox{.}(1991)]%
        {jacobs1991adaptive}
\bibfield{author}{\bibinfo{person}{Robert~A Jacobs}, \bibinfo{person}{Michael~I
  Jordan}, \bibinfo{person}{Steven~J Nowlan}, {and} \bibinfo{person}{Geoffrey~E
  Hinton}.} \bibinfo{year}{1991}\natexlab{}.
\newblock \showarticletitle{Adaptive mixtures of local experts}.
\newblock \bibinfo{journal}{\emph{Neural computation}} \bibinfo{volume}{3},
  \bibinfo{number}{1} (\bibinfo{year}{1991}), \bibinfo{pages}{79--87}.
\newblock


\bibitem[Jordan and Jacobs(1994)]%
        {jordan1994hierarchical}
\bibfield{author}{\bibinfo{person}{Michael~I Jordan} {and}
  \bibinfo{person}{Robert~A Jacobs}.} \bibinfo{year}{1994}\natexlab{}.
\newblock \showarticletitle{Hierarchical mixtures of experts and the EM
  algorithm}.
\newblock \bibinfo{journal}{\emph{Neural computation}} \bibinfo{volume}{6},
  \bibinfo{number}{2} (\bibinfo{year}{1994}), \bibinfo{pages}{181--214}.
\newblock


\bibitem[Ju et~al\mbox{.}(2021)]%
        {ju2021idrlp}
\bibfield{author}{\bibinfo{person}{Mingye Ju}, \bibinfo{person}{Can Ding},
  \bibinfo{person}{Charles~A Guo}, \bibinfo{person}{Wenqi Ren}, {and}
  \bibinfo{person}{Dacheng Tao}.} \bibinfo{year}{2021}\natexlab{}.
\newblock \showarticletitle{IDRLP: Image Dehazing Using Region Line Prior}.
\newblock \bibinfo{journal}{\emph{IEEE Transactions on Image Processing}}
  \bibinfo{volume}{30} (\bibinfo{year}{2021}), \bibinfo{pages}{9043--9057}.
\newblock


\bibitem[Li et~al\mbox{.}(2017a)]%
        {li2017all}
\bibfield{author}{\bibinfo{person}{Boyi Li}, \bibinfo{person}{Xiulian Peng},
  \bibinfo{person}{Zhangyang Wang}, \bibinfo{person}{Jizheng Xu}, {and}
  \bibinfo{person}{Dan Feng}.} \bibinfo{year}{2017}\natexlab{a}.
\newblock \showarticletitle{An all-in-one network for dehazing and beyond}.
\newblock \bibinfo{journal}{\emph{arXiv preprint arXiv:1707.06543}}
  (\bibinfo{year}{2017}).
\newblock


\bibitem[Li et~al\mbox{.}(2017b)]%
        {aod}
\bibfield{author}{\bibinfo{person}{Boyi Li}, \bibinfo{person}{Xiulian Peng},
  \bibinfo{person}{Zhangyang Wang}, \bibinfo{person}{Jizheng Xu}, {and}
  \bibinfo{person}{Dan Feng}.} \bibinfo{year}{2017}\natexlab{b}.
\newblock \showarticletitle{Aod-net: All-in-one dehazing network}. In
  \bibinfo{booktitle}{\emph{Proceedings of the IEEE international conference on
  computer vision}}. \bibinfo{pages}{4770--4778}.
\newblock


\bibitem[Li et~al\mbox{.}(2018a)]%
        {SOTS}
\bibfield{author}{\bibinfo{person}{Boyi Li}, \bibinfo{person}{Wenqi Ren},
  \bibinfo{person}{Dengpan Fu}, \bibinfo{person}{Dacheng Tao},
  \bibinfo{person}{Dan Feng}, \bibinfo{person}{Wenjun Zeng}, {and}
  \bibinfo{person}{Zhangyang Wang}.} \bibinfo{year}{2018}\natexlab{a}.
\newblock \showarticletitle{Benchmarking single-image dehazing and beyond}.
\newblock \bibinfo{journal}{\emph{IEEE Transactions on Image Processing}}
  \bibinfo{volume}{28}, \bibinfo{number}{1} (\bibinfo{year}{2018}),
  \bibinfo{pages}{492--505}.
\newblock


\bibitem[Li et~al\mbox{.}(2020)]%
        {li2020all}
\bibfield{author}{\bibinfo{person}{Ruoteng Li}, \bibinfo{person}{Robby~T Tan},
  {and} \bibinfo{person}{Loong-Fah Cheong}.} \bibinfo{year}{2020}\natexlab{}.
\newblock \showarticletitle{All in one bad weather removal using architectural
  search}. In \bibinfo{booktitle}{\emph{Proceedings of the IEEE/CVF Conference
  on Computer Vision and Pattern Recognition}}. \bibinfo{pages}{3175--3185}.
\newblock


\bibitem[Li et~al\mbox{.}(2018b)]%
        {li2018harmonious}
\bibfield{author}{\bibinfo{person}{Wei Li}, \bibinfo{person}{Xiatian Zhu},
  {and} \bibinfo{person}{Shaogang Gong}.} \bibinfo{year}{2018}\natexlab{b}.
\newblock \showarticletitle{Harmonious attention network for person
  re-identification}. In \bibinfo{booktitle}{\emph{Proceedings of the IEEE
  conference on computer vision and pattern recognition}}.
  \bibinfo{pages}{2285--2294}.
\newblock


\bibitem[Liu et~al\mbox{.}(2019)]%
        {griddehazenet}
\bibfield{author}{\bibinfo{person}{Xiaohong Liu}, \bibinfo{person}{Yongrui Ma},
  \bibinfo{person}{Zhihao Shi}, {and} \bibinfo{person}{Jun Chen}.}
  \bibinfo{year}{2019}\natexlab{}.
\newblock \showarticletitle{Griddehazenet: Attention-based multi-scale network
  for image dehazing}. In \bibinfo{booktitle}{\emph{Proceedings of the IEEE/CVF
  International Conference on Computer Vision}}. \bibinfo{pages}{7314--7323}.
\newblock


\bibitem[Liu et~al\mbox{.}(2021)]%
        {liu2021synthetic}
\bibfield{author}{\bibinfo{person}{Ye Liu}, \bibinfo{person}{Lei Zhu},
  \bibinfo{person}{Shunda Pei}, \bibinfo{person}{Huazhu Fu},
  \bibinfo{person}{Jing Qin}, \bibinfo{person}{Qing Zhang},
  \bibinfo{person}{Liang Wan}, {and} \bibinfo{person}{Wei Feng}.}
  \bibinfo{year}{2021}\natexlab{}.
\newblock \showarticletitle{From Synthetic to Real: Image Dehazing
  Collaborating with Unlabeled Real Data}.
\newblock \bibinfo{journal}{\emph{arXiv preprint arXiv:2108.02934}}
  (\bibinfo{year}{2021}).
\newblock


\bibitem[Liu et~al\mbox{.}(2018)]%
        {liu2018desnownet}
\bibfield{author}{\bibinfo{person}{Yun-Fu Liu}, \bibinfo{person}{Da-Wei Jaw},
  \bibinfo{person}{Shih-Chia Huang}, {and} \bibinfo{person}{Jenq-Neng Hwang}.}
  \bibinfo{year}{2018}\natexlab{}.
\newblock \showarticletitle{DesnowNet: Context-aware deep network for snow
  removal}.
\newblock \bibinfo{journal}{\emph{IEEE Transactions on Image Processing}}
  \bibinfo{volume}{27}, \bibinfo{number}{6} (\bibinfo{year}{2018}),
  \bibinfo{pages}{3064--3073}.
\newblock


\bibitem[Paszke et~al\mbox{.}(2017)]%
        {automatic}
\bibfield{author}{\bibinfo{person}{Adam Paszke}, \bibinfo{person}{Sam Gross},
  \bibinfo{person}{Soumith Chintala}, \bibinfo{person}{Gregory Chanan},
  \bibinfo{person}{Edward Yang}, \bibinfo{person}{Zachary DeVito},
  \bibinfo{person}{Zeming Lin}, \bibinfo{person}{Alban Desmaison},
  \bibinfo{person}{Luca Antiga}, {and} \bibinfo{person}{Adam Lerer}.}
  \bibinfo{year}{2017}\natexlab{}.
\newblock \showarticletitle{Automatic differentiation in pytorch}.
\newblock  (\bibinfo{year}{2017}).
\newblock


\bibitem[Pavlitskaya et~al\mbox{.}(2020)]%
        {pavlitskaya2020using}
\bibfield{author}{\bibinfo{person}{Svetlana Pavlitskaya},
  \bibinfo{person}{Christian Hubschneider}, \bibinfo{person}{Michael Weber},
  \bibinfo{person}{Ruby Moritz}, \bibinfo{person}{Fabian Huger},
  \bibinfo{person}{Peter Schlicht}, {and} \bibinfo{person}{Marius Zollner}.}
  \bibinfo{year}{2020}\natexlab{}.
\newblock \showarticletitle{Using mixture of expert models to gain insights
  into semantic segmentation}. In \bibinfo{booktitle}{\emph{Proceedings of the
  IEEE/CVF Conference on Computer Vision and Pattern Recognition Workshops}}.
  \bibinfo{pages}{342--343}.
\newblock


\bibitem[Pei et~al\mbox{.}(2014)]%
        {pei2014Removing}
\bibfield{author}{\bibinfo{person}{S.~C. Pei}, \bibinfo{person}{Y.~T. Tsai},
  {and} \bibinfo{person}{C.~Y. Lee}.} \bibinfo{year}{2014}\natexlab{}.
\newblock \showarticletitle{Removing rain and snow in a single image using
  saturation and visibility features}. In \bibinfo{booktitle}{\emph{IEEE
  International Conference on Multimedia and Expo Workshops}}.
\newblock


\bibitem[Qin et~al\mbox{.}(2020)]%
        {ffa-net}
\bibfield{author}{\bibinfo{person}{Xu Qin}, \bibinfo{person}{Zhilin Wang},
  \bibinfo{person}{Yuanchao Bai}, \bibinfo{person}{Xiaodong Xie}, {and}
  \bibinfo{person}{Huizhu Jia}.} \bibinfo{year}{2020}\natexlab{}.
\newblock \showarticletitle{FFA-Net: Feature fusion attention network for
  single image dehazing}. In \bibinfo{booktitle}{\emph{Proceedings of the AAAI
  Conference on Artificial Intelligence}}, Vol.~\bibinfo{volume}{34}.
  \bibinfo{pages}{11908--11915}.
\newblock


\bibitem[Rajderkar and Mohod(2013)]%
        {rajderkar2013removing}
\bibfield{author}{\bibinfo{person}{Dhanashree Rajderkar} {and}
  \bibinfo{person}{P.S Mohod}.} \bibinfo{year}{2013}\natexlab{}.
\newblock \showarticletitle{Removing snow from an image via image
  decomposition}. In \bibinfo{booktitle}{\emph{2013 International Conference on
  Emerging Trends in Computing, Communication and Nanotechnology (ICE-CCN)}}.
  \bibinfo{pages}{304–--307}.
\newblock


\bibitem[Redmon et~al\mbox{.}(2016)]%
        {redmon2016you}
\bibfield{author}{\bibinfo{person}{Joseph Redmon}, \bibinfo{person}{Santosh
  Divvala}, \bibinfo{person}{Ross Girshick}, {and} \bibinfo{person}{Ali
  Farhadi}.} \bibinfo{year}{2016}\natexlab{}.
\newblock \showarticletitle{You only look once: Unified, real-time object
  detection}. In \bibinfo{booktitle}{\emph{Proceedings of the IEEE conference
  on computer vision and pattern recognition}}. \bibinfo{pages}{779--788}.
\newblock


\bibitem[Schiapparelli et~al\mbox{.}(2022)]%
        {schiapparelli2022proteomic}
\bibfield{author}{\bibinfo{person}{Lucio~M Schiapparelli},
  \bibinfo{person}{Pranav Sharma}, \bibinfo{person}{Hai-Yan He},
  \bibinfo{person}{Jianli Li}, \bibinfo{person}{Sahil~H Shah},
  \bibinfo{person}{Daniel~B McClatchy}, \bibinfo{person}{Yuanhui Ma},
  \bibinfo{person}{Han-Hsuan Liu}, \bibinfo{person}{Jeffrey~L Goldberg},
  \bibinfo{person}{John~R Yates~III}, {et~al\mbox{.}}}
  \bibinfo{year}{2022}\natexlab{}.
\newblock \showarticletitle{Proteomic screen reveals diverse protein transport
  between connected neurons in the visual system}.
\newblock \bibinfo{journal}{\emph{Cell Reports}} \bibinfo{volume}{38},
  \bibinfo{number}{4} (\bibinfo{year}{2022}), \bibinfo{pages}{110287}.
\newblock


\bibitem[Shafiee et~al\mbox{.}(2017)]%
        {shafiee2017fast}
\bibfield{author}{\bibinfo{person}{Mohammad~Javad Shafiee},
  \bibinfo{person}{Brendan Chywl}, \bibinfo{person}{Francis Li}, {and}
  \bibinfo{person}{Alexander Wong}.} \bibinfo{year}{2017}\natexlab{}.
\newblock \showarticletitle{Fast YOLO: A fast you only look once system for
  real-time embedded object detection in video}.
\newblock \bibinfo{journal}{\emph{arXiv preprint arXiv:1709.05943}}
  (\bibinfo{year}{2017}).
\newblock


\bibitem[Shao et~al\mbox{.}(2020)]%
        {shao2020domain}
\bibfield{author}{\bibinfo{person}{Yuanjie Shao}, \bibinfo{person}{Lerenhan
  Li}, \bibinfo{person}{Wenqi Ren}, \bibinfo{person}{Changxin Gao}, {and}
  \bibinfo{person}{Nong Sang}.} \bibinfo{year}{2020}\natexlab{}.
\newblock \showarticletitle{Domain adaptation for image dehazing}. In
  \bibinfo{booktitle}{\emph{Proceedings of the IEEE/CVF conference on computer
  vision and pattern recognition}}. \bibinfo{pages}{2808--2817}.
\newblock


\bibitem[Szegedy et~al\mbox{.}(2013)]%
        {szegedy2013deep}
\bibfield{author}{\bibinfo{person}{Christian Szegedy},
  \bibinfo{person}{Alexander Toshev}, {and} \bibinfo{person}{Dumitru Erhan}.}
  \bibinfo{year}{2013}\natexlab{}.
\newblock \showarticletitle{Deep neural networks for object detection}.
\newblock \bibinfo{journal}{\emph{Advances in neural information processing
  systems}}  \bibinfo{volume}{26} (\bibinfo{year}{2013}).
\newblock


\bibitem[Wang et~al\mbox{.}(2015)]%
        {wang2015single}
\bibfield{author}{\bibinfo{person}{Jin-Bao Wang}, \bibinfo{person}{Ning He},
  \bibinfo{person}{Lu-Lu Zhang}, {and} \bibinfo{person}{Ke Lu}.}
  \bibinfo{year}{2015}\natexlab{}.
\newblock \showarticletitle{Single image dehazing with a physical model and
  dark channel prior}.
\newblock \bibinfo{journal}{\emph{Neurocomputing}}  \bibinfo{volume}{149}
  (\bibinfo{year}{2015}), \bibinfo{pages}{718--728}.
\newblock


\bibitem[Wang et~al\mbox{.}(2017)]%
        {wang2017A}
\bibfield{author}{\bibinfo{person}{Y. Wang}, \bibinfo{person}{S. Liu},
  \bibinfo{person}{C. Chen}, {and} \bibinfo{person}{B. Zeng}.}
  \bibinfo{year}{2017}\natexlab{}.
\newblock \showarticletitle{A Hierarchical Approach for Rain or Snow Removing
  in A Single Color Image}.
\newblock \bibinfo{journal}{\emph{IEEE Transactions on Image Processing}}
  (\bibinfo{year}{2017}), \bibinfo{pages}{3936--3950}.
\newblock


\bibitem[Wu et~al\mbox{.}(2019)]%
        {wu2019deep}
\bibfield{author}{\bibinfo{person}{Di Wu}, \bibinfo{person}{Si-Jia Zheng},
  \bibinfo{person}{Xiao-Ping Zhang}, \bibinfo{person}{Chang-An Yuan},
  \bibinfo{person}{Fei Cheng}, \bibinfo{person}{Yang Zhao},
  \bibinfo{person}{Yong-Jun Lin}, \bibinfo{person}{Zhong-Qiu Zhao},
  \bibinfo{person}{Yong-Li Jiang}, {and} \bibinfo{person}{De-Shuang Huang}.}
  \bibinfo{year}{2019}\natexlab{}.
\newblock \showarticletitle{Deep learning-based methods for person
  re-identification: A comprehensive review}.
\newblock \bibinfo{journal}{\emph{Neurocomputing}}  \bibinfo{volume}{337}
  (\bibinfo{year}{2019}), \bibinfo{pages}{354--371}.
\newblock


\bibitem[Wu et~al\mbox{.}(2021)]%
        {wu2021contrastive}
\bibfield{author}{\bibinfo{person}{Haiyan Wu}, \bibinfo{person}{Yanyun Qu},
  \bibinfo{person}{Shaohui Lin}, \bibinfo{person}{Jian Zhou},
  \bibinfo{person}{Ruizhi Qiao}, \bibinfo{person}{Zhizhong Zhang},
  \bibinfo{person}{Yuan Xie}, {and} \bibinfo{person}{Lizhuang Ma}.}
  \bibinfo{year}{2021}\natexlab{}.
\newblock \showarticletitle{Contrastive Learning for Compact Single Image
  Dehazing}. In \bibinfo{booktitle}{\emph{Proceedings of the IEEE/CVF
  Conference on Computer Vision and Pattern Recognition}}.
  \bibinfo{pages}{10551--10560}.
\newblock


\bibitem[Xie et~al\mbox{.}(2010)]%
        {xie2010improved}
\bibfield{author}{\bibinfo{person}{Bin Xie}, \bibinfo{person}{Fan Guo}, {and}
  \bibinfo{person}{Zixing Cai}.} \bibinfo{year}{2010}\natexlab{}.
\newblock \showarticletitle{Improved single image dehazing using dark channel
  prior and multi-scale retinex}. In \bibinfo{booktitle}{\emph{2010
  international conference on intelligent system design and engineering
  application}}, Vol.~\bibinfo{volume}{1}. IEEE, \bibinfo{pages}{848--851}.
\newblock


\bibitem[Xu et~al\mbox{.}(2012a)]%
        {xu2012fast}
\bibfield{author}{\bibinfo{person}{Haoran Xu}, \bibinfo{person}{Jianming Guo},
  \bibinfo{person}{Qing Liu}, {and} \bibinfo{person}{Lingli Ye}.}
  \bibinfo{year}{2012}\natexlab{a}.
\newblock \showarticletitle{Fast image dehazing using improved dark channel
  prior}. In \bibinfo{booktitle}{\emph{2012 IEEE international conference on
  information science and technology}}. IEEE, \bibinfo{pages}{663--667}.
\newblock


\bibitem[Xu et~al\mbox{.}(2012b)]%
        {xu2012An}
\bibfield{author}{\bibinfo{person}{J. Xu}, \bibinfo{person}{W. Zhao},
  \bibinfo{person}{P. Liu}, {and} \bibinfo{person}{X. Tang}.}
  \bibinfo{year}{2012}\natexlab{b}.
\newblock \showarticletitle{An Improved Guidance Image Based Method to Remove
  Rain and Snow in a Single Image}.
\newblock \bibinfo{journal}{\emph{Computer and Information Science}}
  \bibinfo{volume}{5}, \bibinfo{number}{3} (\bibinfo{year}{2012}).
\newblock


\bibitem[Xu et~al\mbox{.}(2012c)]%
        {xu2012removing}
\bibfield{author}{\bibinfo{person}{Jing Xu}, \bibinfo{person}{Wei Zhao},
  \bibinfo{person}{Peng Liu}, {and} \bibinfo{person}{Xinaglong Tang}.}
  \bibinfo{year}{2012}\natexlab{c}.
\newblock \showarticletitle{Removing rain and snow in a single image using
  guided filter}. In \bibinfo{booktitle}{\emph{2012 IEEE International
  Conference on Computer Science and Automation Engineering (CSAE)}}.
  \bibinfo{pages}{304–--307}.
\newblock


\bibitem[Zheng et~al\mbox{.}(2011)]%
        {zheng2011person}
\bibfield{author}{\bibinfo{person}{Wei-Shi Zheng}, \bibinfo{person}{Shaogang
  Gong}, {and} \bibinfo{person}{Tao Xiang}.} \bibinfo{year}{2011}\natexlab{}.
\newblock \showarticletitle{Person re-identification by probabilistic relative
  distance comparison}. In \bibinfo{booktitle}{\emph{CVPR 2011}}. IEEE,
  \bibinfo{pages}{649--656}.
\newblock


\bibitem[Zheng et~al\mbox{.}(2013)]%
        {zheng2013single}
\bibfield{author}{\bibinfo{person}{Xianhui Zheng}, \bibinfo{person}{Yinghao
  Liao}, \bibinfo{person}{Wei Guo}, \bibinfo{person}{Xueyang Fu}, {and}
  \bibinfo{person}{Xinghao Ding}.} \bibinfo{year}{2013}\natexlab{}.
\newblock \showarticletitle{Single-image-based rain and snow removal using
  multi-guided filter}. In \bibinfo{booktitle}{\emph{International Conference
  on Neural Information Processing}}. Springer, \bibinfo{pages}{258--265}.
\newblock


\bibitem[Zhu et~al\mbox{.}(2015)]%
        {zhu2015fast}
\bibfield{author}{\bibinfo{person}{Qingsong Zhu}, \bibinfo{person}{Jiaming
  Mai}, {and} \bibinfo{person}{Ling Shao}.} \bibinfo{year}{2015}\natexlab{}.
\newblock \showarticletitle{A fast single image haze removal algorithm using
  color attenuation prior}.
\newblock \bibinfo{journal}{\emph{IEEE transactions on image processing}}
  \bibinfo{volume}{24}, \bibinfo{number}{11} (\bibinfo{year}{2015}),
  \bibinfo{pages}{3522--3533}.
\newblock


\end{thebibliography}
\end{document}